%% file: main.tex
\documentclass{article} 
\usepackage{iclr2025_conference,times}

\usepackage{hyperref}
\usepackage{url}

\usepackage{latexsym}
\usepackage{multirow}
\usepackage{subfigure}
\usepackage{amsmath}
\usepackage{booktabs}
\usepackage{amssymb}
\usepackage{microtype}
\usepackage{balance}
\usepackage{inconsolata}
\usepackage{adjustbox}
\usepackage{graphicx}

\usepackage[utf8]{inputenc} 
\usepackage[T1]{fontenc}    
\usepackage{hyperref}       
\usepackage{url}            
\usepackage{booktabs}       
\usepackage{amsfonts}       
\usepackage{nicefrac}       
\usepackage{microtype}      
\usepackage{xcolor}         

\usepackage{multirow}
\usepackage{subfigure}

\usepackage[english]{babel}
\usepackage{moresize}
\usepackage{amsmath}
\usepackage{algorithmic}
\usepackage{balance}
\usepackage{comment}
\usepackage{paralist}
\usepackage{bm}
\usepackage{pgfplots}
\usetikzlibrary{pgfplots.dateplot}

\usepackage{flushend}
\usepackage[english]{babel}
\usepackage{graphicx}

\usepackage{amssymb}
\usepackage{amsfonts}
\usepackage{url}
\usepackage{bbm}
\usepackage{longtable}
\usepackage{rotating}
\usepackage{multirow}
\usepackage{mathrsfs}
\usepackage{enumitem}
\usepackage[linesnumbered,algoruled,boxed,lined]{algorithm2e}
\usepackage{adjustbox}
\usepackage{hyperref}
\usepackage{pgfplots}
\usetikzlibrary{pgfplots.dateplot}
\usepackage{filecontents}
\usepackage{wrapfig}
\usepackage{caption}

\newcommand{\eg}{\textit{e}.\textit{g}.\xspace}

\def\model{GraphAgent\xspace}

\title{GraphAgent: Agentic Graph \vspace{-0.23in}\\ Language Assistant~~~~~~~~~~~~~~~~~~~~~~~~\includegraphics[scale=0.08]{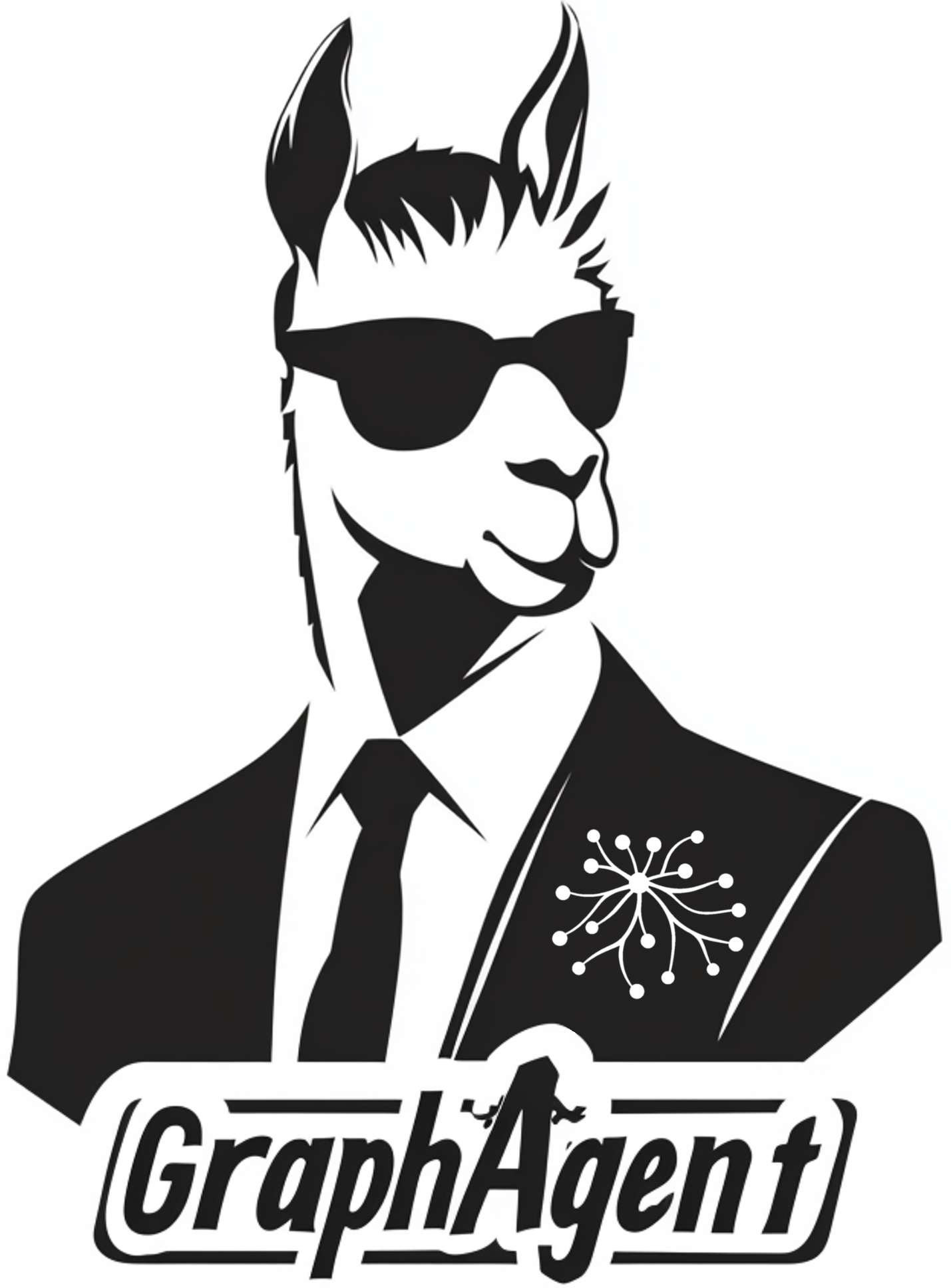}}

\author{Yuhao Yang$^{1}$, Jiabin Tang$^{1}$, Lianghao Xia$^{1}$, Xingchen Zou$^{2}$, Yuxuan Liang$^{2}$, \textbf{Chao Huang}$^{1}$\thanks{Chao Huang is the corresponding author.} \\
University of Hong Kong$^1$ \\
The Hong Kong University of Science and Technology (Guangzhou)$^2$ \\
}

\begin{document}

\maketitle

\begin{abstract}
Real-world data is represented in both structured (\eg, graph connections) and unstructured (\eg, textual, visual information) formats, encompassing complex relationships that include explicit links (such as social connections and user behaviors) and implicit interdependencies among semantic entities, often illustrated through knowledge graphs. In this work, we propose \model, an automated agent pipeline that addresses both explicit graph dependencies and implicit graph-enhanced semantic inter-dependencies, aligning with practical data scenarios for predictive tasks (\eg, node classification) and generative tasks (\eg, text generation). \model comprises three key components: (i) a Graph Generator Agent that builds knowledge graphs to reflect complex semantic dependencies; (ii) a Task Planning Agent that interprets diverse user queries and formulates corresponding tasks through agentic self-planning; and (iii) a Task Execution Agent that efficiently executes planned tasks while automating tool matching and invocation in response to user queries. These agents collaborate seamlessly, integrating language models with graph language models to uncover intricate relational information and data semantic dependencies. Through extensive experiments on various graph-related predictive and text generative tasks on diverse datasets, we demonstrate the effectiveness of our \model\ across various settings. We have made our proposed \model\ open-source at: \textcolor{blue}{\url{https://github.com/HKUDS/GraphAgent}}.
\end{abstract}

\input{intro}
\input{solution}
\input{eval}

\vspace{-0.05in}
\input{relate}

\vspace{-0.05in}
\input{conclusion}

\clearpage

\bibliography{refs}
\bibliographystyle{iclr2025_conference}

\clearpage
\input{appendix}

\end{document}

%% file: intro.tex
\section{Introduction}
\label{sec:intro}

Real-world information exists in a complex ecosystem of interconnected data types. Structured data, particularly graph-based connections, captures explicit relationships such as social networks and user interaction patterns~\citep{fey2023relational}. Complementing this, unstructured data - including text and visual content - reveals implicit semantic relationships between entities~\citep{zhong2023knowledge}. The integration of these diverse data formats has become crucial for modern applications, as it enables more comprehensive and nuanced analysis of complex real-world scenarios~\citep{lu2024llmscore}.

Graph serves as an effective means of representing relational information across various domains. In academic networks, papers are interconnected through explicit citations, with each paper represented as a node in a graph and edges indicating these citations~\cite{chen2023web,wang2022survey}. This structure enables researchers to trace the influence of one paper on another, showcasing the evolution of ideas. Additionally, the papers' content provides unstructured data for analyzing themes, methodologies, and findings. By integrating structured citation data with unstructured text, researchers can identify trends and derive valuable insights, leading to applications such as knowledge summaries and scientific question-answering, which can be framed as \emph{Graph-enhanced Text Generative Tasks}.

In e-commerce scenarios, customer interactions form structured behavior data that can be analyzed in conjunction with unstructured data sources, such as product reviews and descriptions~\cite{shuai2022review,li2023text}. This integrated approach enables businesses to gain deeper insights into consumer behavior patterns and improve recommendation accuracy. Specifically, by integrating user behavior graphs with rich textual information, these user-item interaction forecasting challenges can be effectively approached as \emph{Graph-related Predictive Tasks}.

Existing graph learning methods have become essential frameworks for analyzing and learning from graph data~\citep{hamilton2020graph}. These methods focus on learning embeddings for nodes and edges, mapping structural information into a latent representation space~\citep{yang2020understanding}. Among these, Graph Neural Networks (GNNs) stand out as state-of-the-art (SOTA) approaches~\citep{dai2022towards,liu2022revisiting}. GNNs employ a message-passing mechanism that allows nodes to exchange information with their neighbors, effectively capturing the graph's structural characteristics and enhancing representation learning. However, they primarily focus on explicit graph connections, often neglecting the complex semantic dependencies associated with linked textual data. Additionally, GNNs generally have limited generalization capabilities for real-world graph mining tasks~\citep{xia2024anygraph,mao2024graph}. They often require training task-specific models, which complicates automation and reduces effectiveness in zero-shot scenarios. In practical applications, the ability to process both structured and unstructured data, particularly with unseen new data, is crucial.

\begin{wrapfigure}{R}{0.5\textwidth}
    \vspace{-0.18in}
    \centering
    \includegraphics[width=1.0\linewidth]{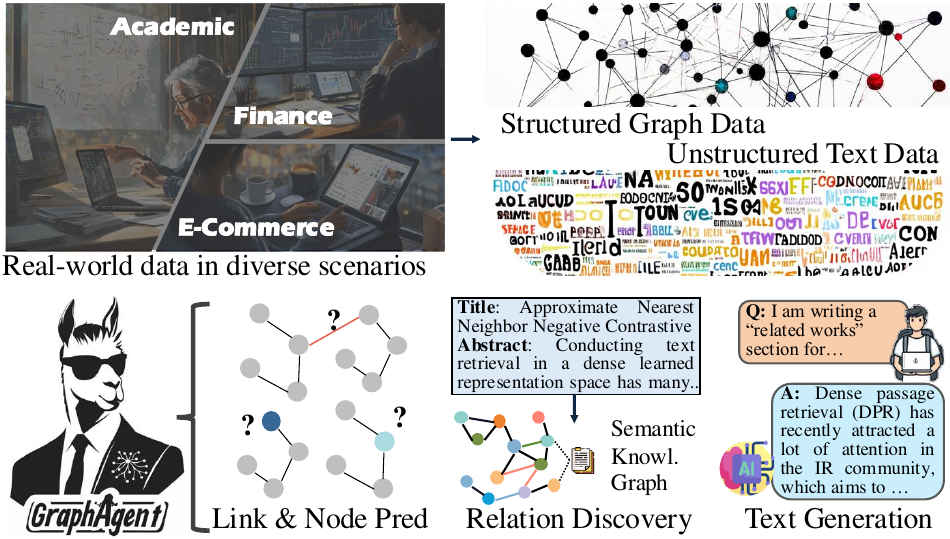}
    \vspace{-0.25in}
    \caption{\model processes both structured and unstructured data, adapting seamlessly to various downstream tasks across diverse scenarios.}
    \vspace{-0.15in}
    \label{fig:intro}
\end{wrapfigure}

Inspired by the recent success of large language models (LLMs), researchers are striving to enhance the generalization capabilities of graph learning models by enabling LLMs to comprehend graph structural information. Notable examples include GraphGPT~\citep{tang2024graphgpt} and LLaGA~\citep{chen2024llaga}, which convert graph-structured data into tokens suitable for LLM input. However, these approaches are primarily designed for conventional graph learning tasks, such as node classification and link prediction. This narrow focus limits their broader application in effectively handling both structured and unstructured data in a more flexible and efficient manner. In light of these limitations, an important question arises: \emph{How can we empower individuals without any background in graph theory or machine learning to analyze their graph data using natural language and obtain the desired predictions and insights?}\\\vspace{-0.12in}

\noindent \textbf{The Presented Work}. In this paper, we aim to establish a fully automated analysis framework capable of handling a wide variety of data types, including both structured and unstructured data. Our framework, \model, is designed to address diverse user needs, encompassing both graph-related predictive and generative tasks. Built on an agentic architecture, \model\ allows users to interact with it using natural language. This intuitive and comprehensive approach thoroughly empowers all individuals to obtain predictions and insights from graph-structured data, tailored to their specific requirements, without requiring specialized knowledge in graph learning.

To achieve our objective, several key challenges must be addressed: i) \textbf{Constructing Potential Semantic Relationships}: How to derive latent semantic connections from complex data. ii) \textbf{Automating Query Understanding and Task Formulation}: How to automatically interpret user query prompts, formulate them into specific tasks (e.g., predictive or generative tasks), and effectively plan those tasks. iii) \textbf{Efficient Task Execution}: How to accurately and effectively implement the formulated tasks and return correct results. To tackle these challenges, our proposed model is designed with an advanced framework comprising three main components: a \textbf{Graph Generator Agent} that constructs Semantic Knowledge Graphs (SKGs) from user text, a \textbf{Task Planning Agent} that interprets queries and formulates tasks, and a \textbf{Graph Action Agent} that automates the task execution. 

To summarize, this work presents the following contributions:

\begin{itemize}[leftmargin=*]


\item \textbf{Complex Practical Data Integration}. Our framework provides robust handling of real-world scenarios by seamlessly merging structured and unstructured data with graph-based entity relationships. This unified approach enables dual capabilities - supporting both predictive analytics and text generation tasks. By allowing natural language interactions, users can directly query and analyze complex data structures, streamlining information extraction and improving accessibility.

\item \textbf{Multi-Agent Workflow}. This work introduces \model, an advanced automated graph language assistant that enhances the integration of structured and unstructured data analysis. It autonomously constructs semantic knowledge graphs (SKGs) from text, formulates predictive and generative tasks from user queries, and efficiently executes these tasks. This seamless collaboration enables GraphAgent to uncover complex relational information and semantic dependencies, significantly improving usability and accessibility in graph analysis.

\item \textbf{Experimental Evaluation}. We validated our model on both structured and unstructured data, showing strong performance across graph predictive tasks and new graph-related text generative tasks. Additionally, we conducted ablation experiments to assess the effectiveness of key modules. It is important to note that our entire agent framework employs relatively small open-source large language models (\eg, LLaMA-8B), yet our model still exhibits significant advantages compared to current state-of-the-art closed-source models (\eg, GPT-4, Gemini) for generation tasks.

\end{itemize}

%% file: solution.tex
\section{Methodology}
\label{sec:solution}

\begin{figure*}
    \centering
    \includegraphics[width=0.92\linewidth]{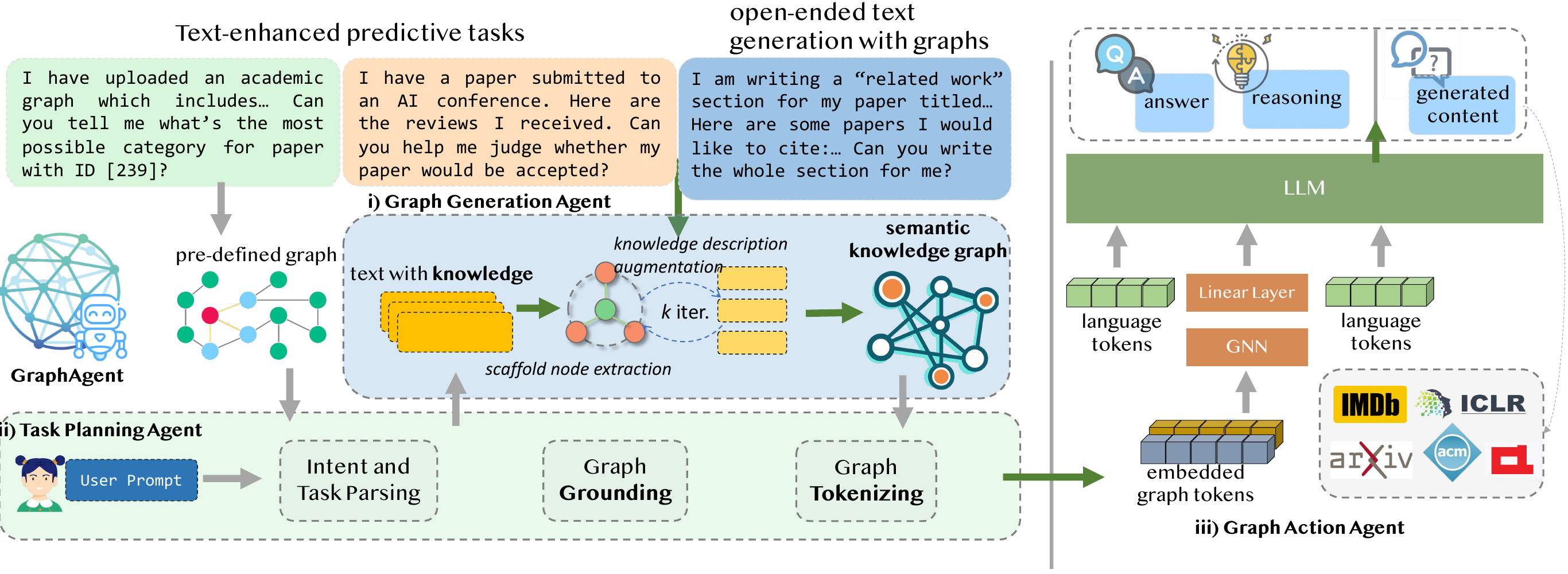}
    \vspace{-0.15in}
    \caption{The overall framework of the proposed \model.}
    \vspace{-0.1in}
    \label{fig:framework}
\end{figure*}

\subsection{Preliminaries}

\noindent \textbf{Graph-empowered Agents}. Our \model\ proposes an automated agentic pipeline that addresses graph predictive and text generation tasks. It can be formulated as $\mathcal{Y} = f(\mathcal{O}; \text{LLM})$, where the agentic function $f(\cdot)$ incorporates an \textbf{Observation} $\mathcal{O}$ that includes both the structured data (\eg, explicit graph connections) or unstructured data (\eg, textual information). The agent then produces an \textbf{Action} $\mathcal{Y}$, which can involve predictions (\eg, node classifications) or text generation tasks (\eg, summarizing text with implicit entity interdependencies). The workflow of \model\ leverages the capabilities of LLMs to enhance its effectiveness in both predictive and generative tasks. \\\vspace{-0.12in}

\noindent \textbf{Graph-Structured Data}. In our \model, both structured and unstructured data are represented as graphs, differing only in the explicitness or implicitness of the entity-wise relationships. To accommodate the diversity of graph data, we utilize heterogeneous graphs to represent the input data. Specifically, a heterogeneous graph is denoted as $\mathcal{G} = (\mathcal{V}, \mathcal{E}, \mathcal{N}, \mathcal{R})$, where $\mathcal{V}$ is the set of all entities, and $\mathcal{E}$ is the set of all edges connecting pairs of entities. The sets $\mathcal{N}$ and $\mathcal{R}$ represent the types of nodes and edges, respectively. For each edge, a meta-type attribute can be retrieved in the form $(n_h, r_i, n_t)$, denoting the meta-types of the head node $n_h$, relation $r_i$, and tail node $n_t$, respectively.

\subsection{Graph Generation Agent}


To uncover the rich contextual information within unstructured data, \model\ designs a Graph Generation Agent that automatically constructs meaningful Semantic Knowledge Graphs (SKGs) from any type of textual input. For example, for a paper abstract that includes the sentence, ``Contrastively trained text-image models have the remarkable ability to perform zero-shot classification'', the model can extract relevant entity nodes such as ``text-image models'' and ``zero-shot classification''. 

\noindent \textbf{Iterative Two-Phase Graph Generation Workflow}. To capture complex implicit entity-wise dependencies, our graph generation agent operates through an automated two-phase workflow: (1) \textbf{Scaffold Knowledge Entity Extraction} and (2) \textbf{Knowledge Description Augmentation}. The first phase is dedicated to identifying key knowledge entities or concepts referred to as scaffold knowledge nodes from the provided text, regardless of its format. Specifically, this phase can be formulated as:
\begin{align}
    \mathcal{V}^{k=0}_{\text{scaffold}} = \text{LLM}(\mathbf{x}_\text{sys\_sk}, \mathbf{g}_s),
\end{align}
where $\mathbf{g}_s$ represents the input unstructured text data, while $\mathbf{x}_{\text{sys\_sk}}$ denotes the system prompt for extracting scaffold knowledge nodes. We adopt an iterative approach to graph generation to capture both high-level and fine-grained semantic dependencies among multi-grained entities. For example, in an academic paper, high-level entities might include "Machine Learning," while fine-grained entities could be "Self-Supervised Learning" and "Graph Neural Network". Specifically, $\mathcal{V}^{k=0}_{\text{scaffold}}$ refers to the generated vertices during the initial iteration ($k = 0$).

The second phase of knowledge augmentation centers on enhancing and enriching the textual descriptions of the generated entity nodes to ensure accurate, comprehensive, and contextually appropriate language modeling. This critical step ensures that each entity is represented with sufficient detail and semantic clarity. Formally, we define this phase as follows:
\begin{align}
    \mathcal{C}^{k=0}_{\text{scaffold}} = \text{LLM}(\mathbf{x}_\text{sys\_ka}, \mathbf{g}_s,  \mathcal{V}^{k=0}_{\text{scaffold}}).
\end{align}
where $\mathcal{C}^{k=0}_{\text{scaffold}}$ denotes the node-specific descriptions, while $\mathbf{x}_{\text{sys\_ka}}$ denotes the system prompt for knowledge augmentation. To iteratively execute this two-phase workflow, \model\ uses the textual augmentation output from the previous round as the implicit graph input for the next round:
\begin{align}
    \mathcal{V}^{k=j}_{\text{scaffold}} &= \text{LLM}(\mathbf{x}_\text{sys\_sk}, \mathcal{C}^{k=j-1}_{\text{scaffold}}) \\
    \mathcal{C}^{k=j}_{\text{scaffold}} &= \text{LLM}(\mathbf{x}_\text{sys\_ka}, \mathcal{C}^{k=j-1}_{\text{scaffold}},  \mathcal{V}^{k=j-1}_{\text{scaffold}}).
\end{align}
We then merge the nodes and descriptions generated across different iterations to form the final node set: $\mathcal{V}_{\text{skg}} = \bigcup \mathcal{V}_{\text{scaffold}}^k$ and $\mathcal{C}_{\text{skg}} = \bigcup \mathcal{C}_{\text{scaffold}}^k$. The relationships among these nodes, denoted as $\mathcal{E}_{\text{skg}}$, are established based on their derivation: if a new node is generated from the textual description of a node in the previous iteration, we connect these two nodes in the semantic knowledge graph. The system prompts used for graph generation are detailed in Table~\ref{tab:sys_prompts}, which is presented in the Appendix.

\subsection{Task Planning Agent}
With both structured and unstructured data represented as graphs, \model\ employs a task planning agent to automatically interpret user queries and transform the graph data into a unified embedding structure. This facilitates easier utilization by the subsequent predictive and generative modules. Input-output examples of the task planning agent is provided in Table~\ref{tab:agent_example} in the Appendix.

\subsubsection{\bf Intent Identification and Task Formulation}
The task planning agent is initially tasked with formulating meaningful predictive or generative tasks based on the user query prompt. Given a user query prompt $\mathbf{x}_{\text{usr\_p}}$ and a predefined system prompt for task parsing $\mathbf{x}_{\text{sys\_tp}}$, the task planning agent formulates the intended task as follows:
\begin{align}
    \mathbf{g}_s, \mathbf{x}_\text{usr\_ann}, \mathbf{t}_\text{usr} = \text{LLM}(\mathbf{x}_\text{sys\_tp}, \mathbf{x}_{\text{usr\_p}}),
\end{align}
This intent identification and task formulation procedure generates three fundamental types of task attributes within our agent architecture, which is specifically defined as follows:
\begin{itemize}[leftmargin=*]
    \item \textbf{Source graph} $\mathbf{g}_s$ represented by formatted files, textual graph descriptions, or plain documents.
    \item \textbf{Task type} $\mathbf{t}_{\text{usr}}$ is inferred from the query prompt and can be one of \textsf{"predictive\_predefined"}, \textsf{"predictive\_wild"}, or \textsf{"open\_generation"}. This task type symbol is used to automatically select system prompt templates during training or inference for different tasks.
    \item \textbf{User annotation} $\mathbf{x}_{\text{usr\_ann}}$ includes additional task information, such as task descriptions, label candidates for predictive tasks, and generation requirements for generative tasks.
\end{itemize}

To construct grounded graph tokens that can be understood by the subsequent action agent, the task planning agent follows two stages: i) Graph-Token Grounding—converting graphs with nodes and edges into grounded Python objects; ii) Graph Tokenization—generating tokens from the input that preserve complex interdependencies among graph-structured entities.

\subsubsection{\bf Graph-Token Grounding}
Our framework reads graph nodes and edges and converts them into grounded Python objects using a graph-building and wrapping tool. Notably, our model can handle diverse graph inputs, regardless of whether an explicit graph with predefined nodes and edges is present. For simplicity, we will demonstrate a scenario where the user uploads a predefined graph. For example, the query prompt might be: "...I want to know which category is correct for the node with ID [305]..." with uploaded graph files such as ["node\_list.txt", "edge\_list.txt"]. To build a grounded graph object in Python, we utilize the graph-building and wrapping tool ($\text{GBW\_Tool}(\cdot)$) with PyG~\citep{fey2019fast} to add nodes and construct edges. Since user-uploaded graphs can have arbitrary node and edge types, we standardize the graphs as heterogeneous graphs, where ${s}_i$ and ${r}_i$ represent the types of each node and edge, respectively. Formally, a heterogeneous graph is constructed as:
\begin{align}
    &\mathcal{G}^{\text{exp}} = \text{GBW\_Tool}(\mathcal{V},\mathcal{E},\mathcal{N},\mathcal{R})\\
    &\mathcal{G}^{\text{skg}} = \text{GBW\_Tool}(\mathcal{V}_{\text{skg}},\mathcal{E}_\text{skg},\mathcal{N}_{\text{skg}},\mathcal{R}_{\text{skg}})
\end{align}
where $\mathcal{V}, \mathcal{E}, \mathcal{N}, \mathcal{R}$ represent the nodes, edges, node types, and edge types of the explicit graph, respectively. They are obtained by parsing the graph input $\mathbf{g}_s$. Similarly,  $\mathcal{V}_\text{skg}, \mathcal{E}_\text{skg}, \mathcal{N}_\text{skg}, \mathcal{R}_\text{skg}$ denote the corresponding graph components generated by the aforementioned Graph Generation Agent. This graph grounding module enables our model to convert graph data from various representations and forms into unified \textsf{Python} objects, facilitating their subsequent utilization.

\subsubsection{\bf Graph Tokenization}
The Task Planning Agent converts discrete nodes and edges into embedded representations suitable for action agents based on graph LLMs. This tokenization process consists of two stages: first, encoding the graph into embeddings, and second, retrieving the nodes and their neighbors to create input graph tokens. For the embedding process, we employ a pre-trained text encoder $f_{\text{text\_enc}}$ and a pre-trained GNN $f_{\text{gnn}}$. Graph tokens are generated by initially encoding the textual features $\mathbf{c}$ of the graph nodes and their meta types using the text encoder, followed by modeling geometric features.
\begin{align}
    \mathbf{e}^{\text{text}}_i &= f_{\text{text\_encoder}}(\mathbf{c}_i); \ \mathbf{e}^{\text{text}}_{{s}_i | {r}_i} = f_{\text{text\_encoder}}(\mathbf{c}_{{s}_i | {r}_i})\\
    \mathbf{e}^{\text{gnn}}_i &= f_{\text{gnn}}(\mathbf{e}^{\text{text}}_i, \mathbf{e}^{\text{text}}_{{s}_i}, \mathbf{e}^{\text{text}}_{{r}_i}, \mathcal{V}, \mathcal{E}).
\end{align}
For each central node $i$ in our heterogeneous graph, we systematically apply a graph sampling tool to create the subgraph input for the subsequent action agent, which can be formulated as follows:
\begin{align}
    [\mathbf{e}^{\text{gnn}}_{N_i}] = \text{Sampling\_Tool}(\mathcal{G}, \mathbf{E}^\text{gnn}, i)
\end{align}

\subsection{Graph Action Agent}
To enhance the capabilities of graph encoding and prediction/generation, we incorporate a trainable Graph Action Agent into our \model\ framework, based on the Graph LLM architecture~\citep{tang2024higpt, chen2024llaga}. This Graph Action Agent is specifically trained to optimize performance for both predictive and text generation tasks involving graph data.

\subsubsection{\bf Cross-Task Graph Agent}
The graph action agent is capable of handling two categories of diverse tasks, as shown below. The details on the system prompt builder and examples of system prompts are shown in Table~\ref{tab:sys_prompts}.
\begin{itemize}[leftmargin=*]
    \item \textbf{Predictive Graph-Language Tasks}. These tasks focus on generating predictions based on user prompts, utilizing both structured and unstructured data. Examples include node classification and link prediction for explicit graph data, as well as document classification based on extracted implicit semantic knowledge graphs (SKGs), such as categorizing news articles. When using implicit SKGs to complement explicit graphs, the graph generator agent uses the observed explicit nodes as initial scaffold nodes to build the SKG. Specifically, for these tasks, our model constructs a system prompt that effectively guides the LLM toward task-specific objectives:
    \begin{align}
        \mathbf{x}_\text{sys\_pred\_i} = f_\text{sys}(\mathbf{t}_\text{usr}, \mathbf{x}_\text{usr\_ann}, \mathbf{g}_s),
    \end{align}
    where the prompt builder function $f_\text{sys}$ creates an appropriate system prompt based on the task type and user annotations, incorporating $\mathbf{g}_s$ for node or graph textual information. The predictive graph-language tasks are then defined as follows:
    \begin{align}
        \mathbf{y}_\text{pred}, \mathbf{y}_{\text{reasoning}} = \text{LLM}(\mathbf{x}_\text{sys\_pred\_i}, \{\mathcal{G}^{\text{exp}}|\mathcal{G}^{\text{skg}}\}),
    \end{align}
    where $\{\mathcal{G}^{\text{exp}}|\mathcal{G}^{\text{skg}}\}$ indicates that the agent can utilize either $\mathcal{G}^{\text{exp}}, \mathcal{G}^{\text{skg}}$ or both. In this context, the LLM generates accurate predictions and reasoning in response to the user's query prompt.\\\vspace{-0.12in}
    
    \item \textbf{Generative Graph-Language Tasks}. The discovered SKGs can serve as robust and comprehensive references for generative language tasks, such as text generation and summarization. These open-ended tasks are typically prompted in a direct text format that implicitly contains knowledge, without the need for predefined graphs. For example, to summarize a news article, an SKG $\mathcal{G}^{\text{skg}}$ is automatically constructed from the article's content, which includes rich entities and connections that aid in the summarization task. Additionally, a system prompt is automatically generated to enhance the content generation quality using the graph-structured information, as follows:
    \begin{align}
        \mathbf{x}_\text{sys\_gen\_i} &= f_\text{sys}(\mathbf{t}_\text{usr}, \mathbf{x}_\text{usr\_ann}, \mathbf{g}_s) \\
        \mathbf{y}_\text{gen} &= \text{LLM}( \mathbf{x}_\text{sys\_gen\_i}, \mathcal{G}^{\text{skg}}),
    \end{align}
    where $\mathbf{y}_\text{gen}$ represents the generated textual output, with input parameters consistent with those used in predictive tasks. In this context, the LLM focuses on producing coherent and contextually accurate content based on both text and graph inputs.
\end{itemize}

\subsubsection{\bf Graph-Instruction Alignment}
\label{sec:alignment}
To teach our agent in comprehending graph-structured data, we implement graph-instruction alignment in the initial fine-tuning stage. Inspired by the work of~\cite{tang2024higpt}, we utilize the efficient, effective, and easily scalable task of graph-instruction matching as our alignment target. Specifically, we present a set of graph token-instruction pairs:
\begin{align}
    \mathcal{D}^{g} = [(\mathbf{e}_0, s_0), (\mathbf{e}_1, s_1), ...];\ \mathcal{D}^c = [(\mathbf{c}_0, \mathbf{c}_{s_0}), (\mathbf{c}_1, \mathbf{c}_{s_1}), ...],
\end{align}
where $(\mathbf{e}_i, s_i)$ denotes the i-th graph token with meta type $s_i$, and $(\mathbf{c}_i, \mathbf{c}_{s_i})$ denotes the text description of the i-th graph token and its meta type, correspondingly. We devise two general tasks to achieve fine-grained and comprehensive alignment between the graph tokens and the textual instructions:
\begin{itemize}[leftmargin=*]
    \item \textbf{Intra-type alignment}. This alignment task aims to strengthen the capability of LLMs to interpret graph embedding tokens of certain meta-types through promoting their alignment with the relevant texts. This is conducted by training LLMs to output correct sequence of the texts given a sequence of graph tokens. Specifically, we construct a dataset $\mathcal{D}^{\text{intra}}$ with each entry consists of two sequences of graph tokens and texts, separately: $d_i^{\text{intra}} = ([(\mathbf{e}_j, s_i), ...], [(\mathbf{c}_k, \mathbf{c}_{s_i}), ...])$. Then, we train the alignment with a next-token-prediction Cross-Entropy objective as follows:
    \begin{align}
        \text{argmin}_\Theta\text{CE\_Loss}(d_i^{\text{intra}}[0]|\text{LLM}(d_i^{\text{intra}}[1])),
    \end{align}
    where $\Theta$ denotes the learnable parameters of the large language model $\text{LLM}(\cdot)$. And indices $[0]$ and $[1]$ indicate the text sequence and the graph token sequence, respectively.\\\vspace{-0.12in}
    
    \item \textbf{Inter-type alignment}. As introducing multiple meta-types in the alignment task can further empower the LLM's comprehension of complex heterogeneous relations, we devise anthor alignment training objective using inter-type graph tokens. Technically, the dataset $\mathcal{D}^{\text{inter}}$ is constructed by sampling entries that consist graph tokens of different meta-types in the first sequence: $d^{\text{inter}}_i=([(\mathbf{e}_m, s_m), (\mathbf{e}_n, s_n), ...], [(\mathbf{c}_n, \mathbf{c}_{s_n}), (\mathbf{c}_q, \mathbf{c}_{s_q}), ...])$. Then, the LLM is trained to predict the text sequence and the meta-type sequence of the provided graph tokens:
    \begin{align}
\text{argmin}_\Theta\text{CE\_Loss}(d^{\text{inter}}_i[0]|d^{\text{inter}}_i[1])).
    \end{align}
\end{itemize}
\vspace{-0.1in}

\subsubsection{\bf Agent Task Finetuning}
\label{sec:multi-task}
To enhance \model's performance on different agent tasks, we propose to finetune the action agent with diverse graph-language instructions covering different agent tasks. Recall that with the task planning agent we have the user requested task $\mathbf{t} \in \mathcal{T}$ from the query prompt. For each $\mathbf{t}$ in the instruction dataset, we pair it with a special systematic prompt to distinguish between various tasks during training. The systematic prompt contains brief description of the task being handled. Formally, the agent task finetuning dataset is constructed as:
\begin{align}
    \mathcal{D}^{multi} = \{(\{(\mathbf{x}_\text{pred}, \mathbf{x}_{\text{reasoning}})|\mathbf{x}_{\text{gen}}\}, \{\mathcal{G}^{\text{exp}}|\mathcal{G}^{\text{skg}}\}, \mathbf{t}_i, \mathbf{a}_i)\},
\end{align}
For each instruction-output pair, the graph provided can be a explicit graph, an automatically discovered SKG, or both. For predictive tasks, the output includes both a prediction and its reasoning, while for generative tasks, the output is the gold-standard objective.

Further, to facilitate a smooth learning curve for multi-tasking the graph language model, we take inspiration from curriculum learning techniques~\citep{xu2020curriculum, bengio2009curriculum} and sort our training tasks into different difficulty levels. We start training with easier tasks to build the model's foundational graph-language understanding. As training progresses, we gradually introduce more complex tasks to refine the model's capabilities. The details are demonstrated in Table~\ref{tab:cur}.

%% file: eval.tex
\section{Evaluation}
\label{sec:eval}

In this section, the effectiveness of our proposed \model\ framework is assessed through a detailed evaluation framework centered around several key Research Questions (\textbf{RQs}):
\begin{itemize}[leftmargin=*]

\item \textbf{RQ1}: How effectively does our \model\ capture both graph relational information and the textual semantic inter-dependencies necessary for graph-related predictive tasks?

\item \textbf{RQ2}: How effective is \model\ at performing predictive tasks by capturing the complex but implicitly textual semantic inter-dependencies preserved within the textual data?

\item \textbf{RQ3}: How does our \model\ perform in graph-enhanced text generation tasks with implicit dependency understanding when compared to state-of-the-art large language models (LLMs)? 

\item \textbf{RQ4}: What effects do the key components of our \model\ framework have on its overall performance, as demonstrated by the ablation studies?

\end{itemize}

\subsection{Experimental Settings}
\subsubsection{\bf Implementation Details}

In our \model\ framework, the task planning agent and the graph generation agent are both powered by \textsf{GPT3.5-Turbo}. We enhance their performance in tackling user queries, planning tasks, and discovering semantic knowledge graphs (SKGs) by incorporating few-shot examples into the system prompts of the large language model (LLM). For graph grounding, we effectively utilize \texttt{PyG} to transform structural information into graph objects. In line with established practices, we employ Sentence-BERT (\textsf{all-mpnet-base-v2}) for text-attributed graph embedding, ensuring a robust semantic representation. For the graph action agent, we build it using Llama3-8b~\cite{dubey2024llama3herdmodels} as the foundational language model. To connect the textual semantic representation space with the graph-structural representation space~\cite{liu2024visual,tang2024graphgpt}, we incorporate a learnable adaptation linear layer. Additionally, we implement a heterogeneous graph model~\cite{tang2024higpt} that has been pre-trained using data from text-graph node pairs. The nodes are encoded with embeddings from the pre-trained model, projected through the learnable adaptation layer, and ultimately processed by the LLM along with relevant language tokens. This integrated approach facilitates seamless interaction between language understanding and graph-based reasoning.

\begin{table*}[t!]
\centering
\caption{Dataset details for training and evaluation. "NC" is short for node classification.}
\label{tab:data}
\vspace{-0.1in}
\resizebox{\linewidth}{!}{
\begin{tabular}{ccccccc}
    \toprule
    & \textbf{IMDB} & \textbf{ACM} & \textbf{Arxiv-Papers} & \textbf{ICLR-Peer Reviews} & \textbf{Related Work Generation} & \textbf{GovReport Summarization} \\
    \midrule
    \textbf{Task Type} & Predictive & Predictive & Predictive & Predictive & Generative & Generative \\
    \textbf{Sub-Task} & NC & NC & Paper Classification & Paper Judgement Prediction & Text Generation & Text Summarization \\
    \textbf{Pre-defined Graph?} & \checkmark & \checkmark & $\times$ & $\times$ & $\times$ & $\times$\\
    \textbf{\#Train Samples} & 2,400 & - & 5,175 & 3,141 & 4,155 & - \\
    \textbf{\#Eval Samples} & - & 1000 & 500 & 500 & 500 & 304 \\
    \textbf{\#Tokens} & 10M & 0.8M & 30M & 45M & 93M & 2M \\
    \textbf{\#Pre-defined Graph Nodes} & 11,616 & 10,942 & - & - & - & - \\
    \textbf{SKG Source} & People Entities & Paper & Paper & Paper, Reviews & Multiple Papers & Documents \\
    \textbf{\#SKG Nodes} & 57,120 & 20,388 & 153,555 & 161,592 & 875,921 & 15,621 \\
    \bottomrule
\end{tabular}
}
\label{tab:dataset_details}
\vspace{-0.1in}
\end{table*}

\subsubsection{\bf Datasets}
To ensure usability across a diverse range of graph agent tasks, we utilize various datasets for evaluating the performance of our \model. A summary of these datasets is provided in Table~\ref{tab:data}. 

\begin{itemize}[leftmargin=*]

\item \textbf{Graph-Related Predictive Tasks}. For tasks that involve explicit graph relational information, we utilize two benchmark datasets: IMDB \cite{fu2020magnn} and ACM \cite{wang2019heterogeneous}. In contrast, for predictive tasks that do not depend on explicit graph structures, we have curated two additional datasets: Arxiv-Papers \cite{he2023harnessing} and ICLR-Peer Reviews\footnote{\url{https://github.com/ranpox/iclr2024-openreview-submissions}}. The Arxiv-Papers dataset comprises published papers from Arxiv in 2023, from which we randomly sampled a subset. This dataset is created by analyzing the titles and abstracts of these papers to classify whether they are likely to be accepted. The ICLR-Peer Reviews dataset features pairs of papers and their corresponding reviews from ICLR 2024, specifically focusing on borderline cases that pose challenges in determining acceptance. This dataset is used for both training and testing purposes.

\item \textbf{Graph-Enhanced Text Generation}. To demonstrate the text generation capabilities of model, we evaluate its performance in generating related work for research papers and summarizing lengthy documents using graph-enhanced semantic dependencies. \textbf{First}, we collected datasets from the ACL and EMNLP conferences, covering the years 2020 to 2023, including both the "main" and "findings" tracks. We extracted the related work sections from these papers and organized them into approximately 5,000 topic-content pairs. For generating related work, \model\ takes a list of paper titles and their corresponding abstracts—input that can be provided by users. Using this information, scaffold knowledge graphs are created and subsequently processed by the Graph Action Agent, which comprehends the data to produce comprehensive related work for the specified papers. \textbf{Second}, we utilize the GovReport dataset~\footnote{\url{https://huggingface.co/datasets/ccdv/govreport-summarization}} to evaluate \model\ as a language assistant for document summarization. This dataset comprises detailed reports from government research agencies, including the Congressional Research Service and the U.S. Government Accountability Office. It necessitates the summarization of longer documents, maintaining richer context and semantic interdependencies, unlike other summarization datasets.



\end{itemize}

\subsubsection{\bf Baseline Methods}
We incorporate a diverse range of baseline models from various research domains to ensure a comprehensive comparison. Specifically, we examine methods for graph-related predictive tasks, including homogeneous GNNs, heterogeneous models, and graph LLMs. Additionally, we utilize and compare state-of-the-art large language models—both open-source and closed-source—alongside retrieval-augmented generation (RAG) systems for enhanced text generation.

\begin{itemize}[leftmargin=*]

\item \textbf{Graph-Related Predictive Tasks}. We consider baseline methods from three key areas: i) \textbf{Homogeneous GNNs}, which include SAGE~\cite{graphsage} and GAT~\cite{gat} as representative models; ii) \textbf{Heterogeneous Graph Models}, featuring the specialists such as HAN~\cite{han}, HGT~\cite{HGT}, and HetGNN~\cite{hetgnn}; and iii) \textbf{Graph LLMs}, for which we adopt HiGPT~\cite{tang2024higpt}, a state-of-the-art heterogeneous graph language model that is particularly well-suited for managing complex heterogeneous structures.

\item \textbf{Graph-Enhanced Text Generation}. We utilize a variety of state-of-the-art large language models (LLMs), categorized as follows: i) \textbf{Open-Source LLMs} include the Llama 3 series~\cite{dubey2024llama3herdmodels}, Mistral NeMo\footnote{https://mistral.ai/news/mistral-nemo/}, and Qwen2-72b~\cite{yang2024qwen2}; ii) \textbf{Closed-Source Commercial LLMs} consist of Deepseek-Chat-V2, GPT4o-mini, and Gemini-1.5-Flash, using their API services for empirical results; iii) \textbf{LLM-empowered RAG Systems}. We also compare \model with GraphRAG\footnote{\url{https://github.com/microsoft/graphrag}}, which enhances LLMs through graph-based retrieval-augmented generation.

\end{itemize}

\subsubsection{\bf Evaluation Protocols}

We implement comprehensive and consistent training strategies across all models. We apply full fine-tuning for our model and all baseline models requiring supervised fine-tuning. For model selection, we utilize validation sets with early-stopping for predictive tasks, while monitoring training loss decreasing rate for alignment training and generative tasks. To ensure fair comparison, we maintain consistent feature encoder (all-mpnet-base-v2) across all models including GNNs and Graph LLMs. We use identical prompt templates across all LLM-based models, with GraphLLMs receiving additional graph tokens for embedding injection and basic meta type descriptions (detailed in Table~\ref{tab:sys_prompts}). The iterative steps are set to 2 for discovering two-hop knowledge graphs per query prompt.

For evaluation, we adopt different metrics based on task types. In graph-related predictive tasks with ground truth, we use Micro-F1 (Mi-F1), Macro-F1 (Ma-F1), and AUC metrics. For graph-enhanced generative tasks that are open-ended, we primarily rely on the PPL score using state-of-the-art models (Llama3-70b, Qwen2-72b) to measure fluency, rather than reference-based similarity metrics which can be misleading due to their limitations in text generation evaluation. Additionally, we incorporate the LLM-as-judge approach for better approximation of human judgment. This comprehensive evaluation framework ensures robust and meaningful comparison across different model architectures while addressing the limitations of conventional evaluation metrics for generative tasks.

\begin{table*}[t]
 \vspace{-0.15in}
 \centering
 \caption{Zero-shot learning performance evaluation: We assess our model's transfer capabilities by training on IMDB dataset with few-shot learning, then evaluating node classification performance on ACM dataset under zero-shot conditions, utilizing both graph structural and textual information.
}\label{tab:performance_1}
 \vspace{-0.1in}
 \setlength{\tabcolsep}{0.6mm}
 \ssmall
 \resizebox{\textwidth}{!}
 {
 \begin{tabular}{cccccccccc}
 \hline
 Metric & Trained on & SAGE & GAT & HAN & HGT & HetGNN &{HiGPT} & \textbf{\model} &\textit{Imprv.} \\
 \hline
 \multirow{1}{*}{Micro-F1}& IMDB-1 & 32.93$\pm$4.18 & 35.67$\pm$0.53 & 34.07$\pm$1.11 & 32.40$\pm$0.14 & 37.43$\pm$4.34 & {45.40$\pm$0.89} & \textbf{51.21$\pm$1.32} & \textit{12.8\%} \\
 (\%) & IMDB-40 & 31.73$\pm$0.05 & 23.93$\pm$1.44 & 26.97$\pm$1.94 & 35.60$\pm$0.99 & 31.80$\pm$0.16 & {50.50$\pm$0.77} & \textbf{74.98$\pm$1.24} & \textit{48.5\%}\\
 \cline{1-10}
 \multirow{1}{*}{Macro-F1}& IMDB-1 & 26.47$\pm$2.69 & 29.08$\pm$1.31 & 22.50$\pm$4.16 & 16.31$\pm$0.05 & 31.39$\pm$4.68 & {41.77$\pm$1.24} & \textbf{46.82$\pm$1.43} & \textit{12.1\%}\\
 (\%)& IMDB-40 & 31.17$\pm$0.17 & 21.41$\pm$0.71 & 23.13$\pm$1.32 & 27.49$\pm$1.22 & 31.44$\pm$0.17 & {45.85$\pm$0.89} & \textbf{74.98$\pm$1.12} & \textit{63.5\%}\\
 \cline{1-10}
 \multirow{1}{*}{AUC}& IMDB-1 & 49.34$\pm$2.47 & 52.48$\pm$0.38 & 51.28$\pm$0.86 & 50.00$\pm$0.00 & 53.18$\pm$2.95 & {59.69$\pm$0.82} & \textbf{64.10$\pm$1.25} & \textit{7.4\%}\\
 (\%) & IMDB-40 & 48.67$\pm$0.13 & 43.20$\pm$1.08 & 45.45$\pm$1.46 & 51.48$\pm$0.43 & 48.72$\pm$0.06 & {63.60$\pm$0.51} & \textbf{80.90$\pm$1.01} & \textit{27.2\%}\\
 \hline
 \end{tabular}
 }
 \vspace{-0.2in}
 \end{table*}

\subsection{Graph Prediction Task with Explicit and Implicit Graph Contexts (RQ1)}\vspace{-0.05in}
We investigate \model's performance on graph-related prediction tasks, specifically node classification with explicit graph structures. Our approach enhances existing methods by automatically incorporating a semantic knowledge graph from node text, utilizing both the semantic KG and explicit graph connections as dual sources for graph token input. Following recent works~\cite{tang2024graphgpt,tang2024higpt,chen2024llaga}, we employ a fully zero-shot evaluation framework to better assess real-world applicability. Our experimental setup involves training models on the IMDB dataset under few-shot settings (1 shot and 40 shots), then evaluating performance on 1,000 previously unseen nodes from the ACM dataset. For our method and other LLM-enhanced approaches, we incorporate Chain-of-Thought~\cite{wei2022chain} for inference augmentation.

\begin{wrapfigure}{R}{0.5\textwidth}
\centering
\caption{Performance comparison with state-of-the-art LLMs on complex graph prediction tasks involving implicit semantic relationships. Results marked with * indicate statistical significance (p<0.01) compared to the second-best performer.}
\vspace{-0.1in}
\label{tab:predictive}
\setlength{\tabcolsep}{0.5mm}
\resizebox{\linewidth}{!}{
\begin{tabular}{lccccccc}
\toprule
\multirow{2}{*}{Method} & \multirow{2}{*}{Model Size} & \multicolumn{3}{c}{Arxiv-Papers} & \multicolumn{3}{c}{ICLR-Peer Reviews} \\
\cmidrule(lr){3-5}\cmidrule(lr){6-8}
~ & & Mi-F1 & Ma-F1 & AUC & Mi-F1 & Ma-F1 & AUC \\
\midrule
\multicolumn{8}{c}{\textbf{Open-sourced LLMs}} \\
\cmidrule(lr){1-8}
Llama3-8b & 8B & 0.514 & 0.289 & 0.527 & 0.402 & 0.394 & 0.502  \\
Mistral-Nemo & 12B & 0.510 & 0.292 & 0.615 & 0.272 & 0.246 & 0.380  \\
Llama3-70b & 70B & 0.630 & 0.330 & 0.635 & 0.434 & 0.421 & 0.551 \\
Qwen2-72b & 72B & 0.632 & 0.472 & 0.700 & 0.344 & 0.277 & 0.509 \\
\midrule
\multicolumn{8}{c}{\textbf{API-based Commercial LLMs}} \\
\cmidrule(lr){1-8}
Deepseek-Chat-V2 & 236B$\rightarrow$21B & 0.746 & 0.580 & 0.757 & 0.362 & 0.312 & 0.516 \\
GPT4o-mini & - & 0.592 & 0.343 & 0.634 & \textbf{0.692}$^\ast$ & 0.592 & 0.591 \\
Gemini-1.5-Flash & - &  0.748 & 0.504 & 0.714 & 0.684 & 0.487 & 0.533  \\
\midrule
\multicolumn{8}{c}{\textbf{Finetuned LLMs}} \\
\cmidrule(lr){1-8}
Llama3-8b Finetuned & 8B & 0.794 & 0.593 & 0.736 & 0.620 & 0.554 & 0.553  \\
\midrule
\multicolumn{8}{c}{\textbf{GraphRAG Implementations}} \\
\cmidrule(lr){1-8}
Llama3-8b + GraphRAG & 8B & 0.516 & 0.288 & 0.601 & 0.430 & 0.427 & 0.517 \\
Llama3-70b + GraphRAG & 70B & 0.603 & 0.324 & 0.623 & 0.308 & 0.296 & 0.401  \\
\midrule
\textbf{\model-Task Expert}\ & 8B & 0.820 & 0.620 & 0.768 & {0.686} & \textbf{0.620}$^\ast$ & \textbf{0.615}$^\ast$  \\
\textbf{\model-General}\ & 8B & \textbf{0.840}$^\ast$ & \textbf{0.621}$^\ast$ & \textbf{0.769}$^\ast$ & 0.667 & 0.604 & 0.607 \\
\textbf{\model-Zero-Shot}\ & 8B & 0.739 & 0.512 & 0.701 & 0.538 & 0.531 & 0.563\\

\bottomrule
\end{tabular}
}
\vspace{-0.18in}
\end{wrapfigure}

The results summarized in Table~\ref{tab:performance_1} demonstrate that our agent-based approach \model significantly advances the state-of-the-art in predictive graph tasks. Specifically, \model achieves an average improvement of over 28\% across all metrics compared to the previous state-of-the-art graph language model, HiGPT. These substantial improvements stem from the synergistic integration of several key components: a graph generation agent, an automated task planning agent, and dual fine-tuning mechanisms (graph-text alignment and agent task fine-tuning). Together, these components enable \model to excel at constructing rich semantic knowledge graphs, capturing comprehensive inter-dependencies, and understanding complex relationships in both structured and unstructured graph contexts. This architecture translates into superior performance across downstream tasks.

\subsection{Graph Prediction with Implicit Semantic Interdependencies (RQ2)}
We evaluate \model's effectiveness on predictive tasks that require understanding complex semantic interdependencies, comparing against state-of-the-art LLMs. For these tasks, \model\ constructs semantic knowledge graphs (SKGs) by extracting implicit relational patterns through its dual-agent system of task planning and graph generation. The resulting SKG nodes act as semantic anchors, enriching the input representation through embedded and tokenized forms. Our empirical evaluation on Arxiv-Papers and ICLR-Peer Reviews datasets (Table~\ref{tab:predictive}) demonstrates \model's capabilities across three configurations: task-specific (\model-Task Expert), comprehensive (\model-General), and zero-shot generalization (\model-Zero-Shot). Unlike conventional GNNs and GraphLLMs that require explicit graph structures, \model competes directly with leading LLMs of various scales, including fine-tuned and GraphRAG-augmented variants. The experimental results reveal three distinct advantages of our approach:

\begin{itemize}[leftmargin=*]

\item \textbf{Superior Performance with Smaller Model Size.} Despite having only 8B parameters, \model consistently outperforms larger LLMs, including Llama3-70b and Qwen2-72b, achieving a 31.9\% improvement across all metrics on both datasets. By explicitly capturing complex interdependencies via semantic graph structures while maintaining contextual awareness across different semantic levels, \model effectively integrates both local and global information patterns. This architectural approach enables robust handling of intricate reasoning tasks, where both detailed semantic relationships and broader contextual coherence are crucial for accurate predictions. \\\vspace{-0.12in}

\item \textbf{Robust Generalization Through Multi-task and Zero-shot Learning.} \model exhibits exceptional adaptability and robust performance across different learning scenarios. The multi-task variant, \model-General, demonstrates superior performance compared to task-specific models on Arxiv-Papers, showcasing enhanced comprehension and reasoning capabilities over text-graph pairs through self-constructed SKGs. While there is a modest performance trade-off on ICLR-Peer Reviews, the multi-task model maintains competitive results comparable to specialized versions. Notably, \model shows impressive zero-shot generalization: even with domain transfer challenges, our 8B model achieves performance parity with state-of-the-art LLMs like Deepseek-Chat-V2 and Gemini-1.5-Flash. These findings demonstrate how our approach of integrating semantic knowledge graphs and specialized tuning techniques can significantly enhance model capabilities through structured knowledge representation. \\\vspace{-0.12in}

\item \textbf{Superior Performance over Vanilla SFT and GraphRAG.} Comparative experiments demonstrate \model's significant advantages over both vanilla supervise fine-tuning (SFT) LLMs and GraphRAG implementations. This performance gain can be attributed to two key factors: First, compared to vanilla supervised fine-tuning SFT LLMs, \model effectively leverages the LLM's knowledge base through our semantic KG integration paradigm, leading to enhanced performance. Second, while GraphRAG uses the same knowledge references, \model's graph embedding token approach provides a more efficient and consolidated knowledge representation. This not only reduces input token overhead but also helps mitigate LLM hallucination through structured knowledge encoding, ultimately resulting in more reliable and robust performance.

\end{itemize}

\begin{wrapfigure}{R}{0.5\textwidth}
\centering
\caption{Performances on ACL-EMNLP related works content generation. Light grey denotes that the score is computed with the same-family model.}
\vspace{-0.1in}
\label{tab:nlp_rw}
\resizebox{\linewidth}{!}{
\begin{tabular}{lccccc}
\toprule
\multirow{2}{*}{Method} & \multirow{2}{*}{Model Size} & \multicolumn{2}{c}{PPL-Llama3-70b} & \multicolumn{2}{c}{PPL-Qwen2-72b} \\
\cmidrule(lr){3-4}\cmidrule(lr){5-6}
~ & & Mean & Max & Mean & Max \\
\midrule
\multicolumn{6}{c}{\textbf{Open-sourced LLMs}} \\
\cmidrule(lr){1-6}
Llama3-8b & 8B & \colorbox{gray!30}{7.016} & \colorbox{gray!30}{13.061} & 7.491 & 12.787 \\
Mistral-Nemo & 12B & 7.367 & 15.967 & 6.872 & 12.065 \\
Llama3-70b & 70B & \colorbox{gray!30}{6.168} & \colorbox{gray!30}{14.436} & 5.877 & 12.897 \\
Qwen2-72b & 72B & 6.043 & 11.675 & \colorbox{gray!30}{5.325} & \colorbox{gray!30}{11.302}\\
\midrule
\multicolumn{6}{c}{\textbf{API-based Commercial LLMs}} \\
\cmidrule(lr){1-6}
Deepseek-Chat-V2 & 236B$\rightarrow$21B & 5.632 & 13.483 & 5.144 & 10.337 \\
GPT4o-mini & - & 7.277 & 15.480 & 6.818 & 13.267\\
Gemini-1.5-Flash & - & 5.188 & 10.399 & 5.377 & 10.779 \\
\midrule
\multicolumn{6}{c}{\textbf{Finetuned LLMs}} \\
\cmidrule(lr){1-6}
Llama3-8b Finetuned & 8B & 7.682 & 19.452 & 7.629 & 18.757\\
\midrule
\multicolumn{6}{c}{\textbf{GraphRAG Implementations}} \\
\cmidrule(lr){1-6}
Llama3-8b + GraphRAG & 8B & 7.098 & 18.092 & 6.539 & 14.722 \\
Llama3-70b + GraphRAG & 70B & 6.590 & 14.827 & 6.135 & 14.163  \\
\midrule
\textbf{\model-Task Expert}\ & 8B & 3.805 & 10.316 &4.069 & 11.685\\
\textbf{\model-General}\ & 8B & \textbf{3.618}$^\ast$ & \textbf{8.000}$^\ast$ & \textbf{3.867}$^\ast$ & \textbf{8.775}$^\ast$\\
\bottomrule
\end{tabular}
}
\vspace{-0.18in}
\end{wrapfigure}

\subsection{Graph-Enhanced Text Generation (RQ3)}
\begin{wrapfigure}{R}{0.5\textwidth}
    \centering
    \includegraphics[width=0.94\linewidth]{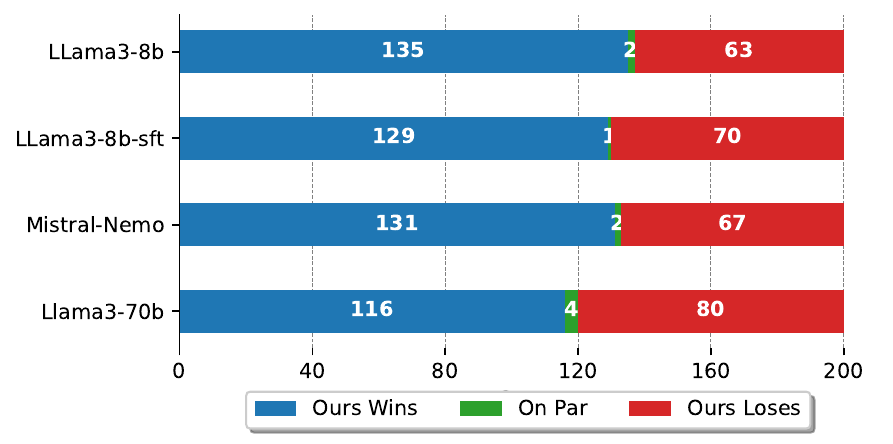}
    \vspace{-0.1in}
    \caption{Comparative evaluation results: GPT-4o as judge assessing our proposed \model\ framework against state-of-the-art open-source LLMs.}
    \label{fig:llm_as_judge}
    \vspace{-0.2in}
\end{wrapfigure}

We evaluate \model's performance on graph-enhanced text generation tasks using both perplexity (PPL) metrics and LLM-based assessment. Results for our evaluated text generation tasks are presented in Table~\ref{tab:nlp_rw} and Figure~\ref{fig:llm_as_judge}, while zero-shot generalization results on GovReport data are shown in Table~\ref{tab:govreport}.

$\bullet$ \textbf{Enhanced Generation Quality via Lower Perplexity}. Table~\ref{tab:nlp_rw} demonstrates \model's superior performance with lower perplexity scores as compared to baselines, as validated by both Llama3-70b and Qwen2-72b. The generated content exhibits enhanced fluency and clarity compared to larger LLMs. We observe that both SFT and GraphRAG variants show performance degradation, indicating that neither simple input-output fine-tuning nor direct knowledge injection through prompts can effectively capture the complex reasoning patterns required for understanding intricate contextual relationships. In contrast, our approach leverages automatically constructed semantic knowledge graphs to substantially enhance the model's reasoning and comprehension capabilities.

$\bullet$ \textbf{Superior Generation Quality via LLM-based Evaluation.} To rigorously validate our model's alignment with human preferences for the generated content, we employed the LLM-as-judge methodology~\cite{zheng2024judging}, which demonstrates stronger correlation with human judgment compared to traditional metrics like BLEU~\cite{papineni2002bleu} and ROUGE~\cite{lin2004rouge}. Using GPT-4 as the judge (evaluation prompts detailed in Table~\ref{tab:llm_as_judge_prompt}), we compared \model against several strong baselines: Llama3-8b, Llama3-8b fine-tuned, Mistral Nemo, and Llama3-70b.

Evaluation on 200 samples from the text generation test set (Figure~\ref{fig:llm_as_judge}) demonstrates \model's superior performance: achieving 114\% quality improvement over Llama3-8b and 45\% over Llama3-70b. \model generates higher quality content in 67\% of cases compared to same-sized models and outperforms leading open-source models in 58\% of instances, despite having only 8B parameters and requiring minimal additional input overhead. These results validate our \model's effectiveness in leveraging semantic knowledge graphs for enhanced text generation capabilities.

\begin{wrapfigure}{R}{0.5\textwidth}
\centering
\caption{GovReport summarization performance. Evaluation scores are presented with same-family model comparisons highlighted in light grey.}
\vspace{-0.1in}
\label{tab:govreport}
\resizebox{\linewidth}{!}{
\begin{tabular}{lccccc}
\toprule
\multirow{2}{*}{Method} & \multirow{2}{*}{Model Size} & \multicolumn{2}{c}{PPL-Llama3-70b} & \multicolumn{2}{c}{PPL-Qwen2-72b} \\
\cmidrule(lr){3-4}\cmidrule(lr){5-6}
~ & & Mean & Max & Mean & Max \\
\midrule

Llama3-8b & 8B & \colorbox{gray!30}{9.476} & \colorbox{gray!30}{25.355} & 7.564 & 17.443 \\
Mistral-Nemo & 12B & 9.333 & 28.537 & 7.194 & 19.347 \\
Llama3-70b & 70B & \colorbox{gray!30}{6.473} & \colorbox{gray!30}{14.724} & 5.629 & 11.813 \\
Qwen2-72b & 72B & 7.134 & 16.075 & \colorbox{gray!30}{5.494} & \colorbox{gray!30}{11.294}\\

Deepseek-Chat-V2 & 236B$\rightarrow$21B & 8.246 & 21.176 & 7.311 & 18.092 \\
GPT4o-mini & - & 10.332 & 23.300 & 6.576 & 10.213 \\
Gemini-1.5-Flash & - & 7.374 & 18.408 & 6.133 & 9.237 \\
\midrule
\textbf{\model-General}\ & 8B & 6.736 & 20.362 & 5.936 & 27.196 \\
\bottomrule
\end{tabular}
}
\vspace{-0.2in}
\end{wrapfigure}

$\bullet$ \textbf{Cross-domain Performance on Document Summarization.} The effectiveness of \model\ extends beyond academic writing to document summarization tasks, as demonstrated in our graph-enhanced text generation evaluation on GovReport data (Table~\ref{tab:govreport_exp_1} shown in Appendix). Notably, without any task-specific optimization, \model exhibits strong structural reasoning abilities by generating well-organized summaries (highlighted in \colorbox{green!30}{green}). This successful transfer of capabilities across domains underscores the model's robust generalization potential. Experimental results from Table~\ref{tab:govreport} shown in Appendix demonstrate \model's competitive performance in zero-shot generative tasks with graphs. The model achieves significantly lower perplexity (PPL) scores compared to same-sized counterparts like Llama3-8b and even the larger Mistral-Nemo. Moreover, \model matches the fluency levels of leading closed-source and open-source LLMs in generating GovReport summaries. These findings suggest that our approach of automatically extracting and leveraging semantic knowledge graphs from input content, combined with diverse multi-task graph-based training, enables robust zero-shot performance.

\vspace{-0.2in}
\subsection{Qualitative Analysis of Graph-enhanced Text Generation Tasks} 
We evaluated \model against Llama3-8b and Llama3-70b on two distinct graph-enhanced text generation tasks, with results presented in Tables~\ref{tab:nlp_rw_exp_1} and~\ref{tab:govreport_exp_1} (Appendix). The experiments demonstrate \model's significant performance advantages over Llama3-8b while achieving comparable results to the much larger Llama3-70b. Notably, in academic writing tasks (Table~\ref{tab:nlp_rw_exp_1}), \model effectively leverages knowledge graphs to capture citation relationships and research development paths, producing well-organized summaries (highlighted in \colorbox{green!30}{green}). In contrast, Llama3-8b exhibits notable limitations in both instruction following and citation formatting accuracy (highlighted in \colorbox{red!30}{red}).

This section presents our automatically generated semantic knowledge graphs (SKGs) through two visualized examples in Tables~\ref{tab:skg_example_1} and~\ref{tab:skg_example_2} from GovReport and Arxiv datasets. We visualize each SKG at two levels: $k=0$ hop showing high-level aspect nodes (highlighted in \colorbox{green!30}{green}) and $k=1$ hop displaying keyword nodes (highlighted in \colorbox{blue!30}{blue}), along with augmented textual attributes for $k=0$ scaffold nodes. These examples demonstrate our Graph Generation Agent's capability in extracting crucial information, revealing multi-hop relationships, and generating coherent semantic representations. For the related work text generation, while containing paper-based SKGs as multiple sub-graphs, individual papers follow similar SKG patterns to the Arxiv examples.

Table~\ref{tab:agent_example} demonstrates the multi-agent collaboration workflow using the IMDB movie classification dataset as an example. The process begins when the \emph{Task Planning Agent} receives a user request, performing request decomposition and task planning, while also processing graph data in scenarios with pre-defined graphs. Next, the \emph{Graph Generation Agent} constructs a task-relevant semantic knowledge graph. Finally, the \emph{Multi-Task Agent} automatically constructs a system prompt based on the task type and executes task inference. This workflow highlights \model's versatility in automatically adapting to different scenarios, whether handling pre-defined graphs or generating new ones, while maintaining consistent performance across various task types.

\begin{table*}
    \centering
    \footnotesize
    \caption{Illustrative Examples of \model's Multi-Agent Pipeline.}
    \vspace{-0.1in}
    \label{tab:agent_example}
    \begin{tabular}{p{1.\textwidth}}
    \toprule
    \midrule
    \textbf{[User]} \\
    Here I have uploaded a relational graph involving movies, directors and actors. Can you tell me which category does the movie with node \colorbox{red!30}{<GRAPH\_NODE\_ID\_[7]>} belong to? Is it action, comedy or drama? \colorbox{red!30}{[Predictive Task with Pre-defined Graph Relations]} \\
    \midrule
    \textbf{[Agent (Task Planning Agent)]} \\
    Sure! I will first decompose your request as follows for further actions: \newline \colorbox{blue!30}{\parbox{\dimexpr\linewidth-2\fboxsep\relax}{<graph source: [uploaded files]>, <task type: predictive\_predefined>, <user annotation>: <"Can you tell me which category does the movie with the target node? Is it action, comedy or drama?">}}  \\
    \textbf{[Agent (Task Planning Agent)]} \\
    Since pre-defined graphs are given, I will try to ground the edge and node list files into readable objects. Then I will encode the graph and index the neighbors for node ID [7] for further actions. 
    \newline
    \colorbox{green!30}{[\textbf{Function Call}: <Graph Grounding> GBW\_Tool($\cdot$)]}
    $\rightarrow$
    \colorbox{blue!30}{$\mathcal{G}^\text{exp}, \mathbf{E}^\text{gnn}$}
    \\
    \textbf{[Agent (Graph Generation Agent)]} \\
    Now the graph tokens for the pre-defined graph are ready. I will further try to discover a semantic knowledge graph for the target node and the sampled neighbors.
    \newline
    \colorbox{green!30}{[Detecting and extracting scaffold nodes for the texts of the nodes...]} $\rightarrow$     \colorbox{blue!30}{$\mathcal{V}^{k=0}_{\text{scaffold}}$}
    \newline
    \colorbox{green!30}{[Augmenting textual descriptions for the scaffold nodes]} $\rightarrow$     \colorbox{blue!30}{$\mathcal{C}^{k=0}_{\text{scaffold}}$}
    \newline
    \colorbox{green!30}{[Iteratively performing until a two-hop knowledge graph is constructed.]}
    \newline
    \colorbox{green!30}{[\textbf{Function Call}: GBW\_Tool($\cdot$)]}
    $\rightarrow$
    \colorbox{blue!30}{$\mathcal{G}^\text{skg}, \mathbf{E}^\text{gnn\_skg}$}\\
    \textbf{[Agent (Graph Action Agent)]} \\
    Since we already have the pre-defined graph and the built SKG, I will construct an appropriate system prompt according to the user request.
    \newline
    \colorbox{blue!30}{<$\mathbf{x}_\text{system\_pred}$>: You are a powerful assistant in accomplishing diverse user required tasks with the help of graphs...}\\
    \textbf{[Agent (Graph Action Agent)]} \\
    \colorbox{blue!30}{\parbox{\dimexpr\linewidth-2\fboxsep\relax}{According to the provided information, the neighbor nodes and the semantic knowledge graph nodes, the most likely category for the movie is ... The reasoning process behind is...}}\\

    \bottomrule
    \end{tabular}
    \vspace{-0.15in}
\end{table*}

\begin{wrapfigure}{R}{0.5\textwidth}
    \centering
    \includegraphics[width=\linewidth]{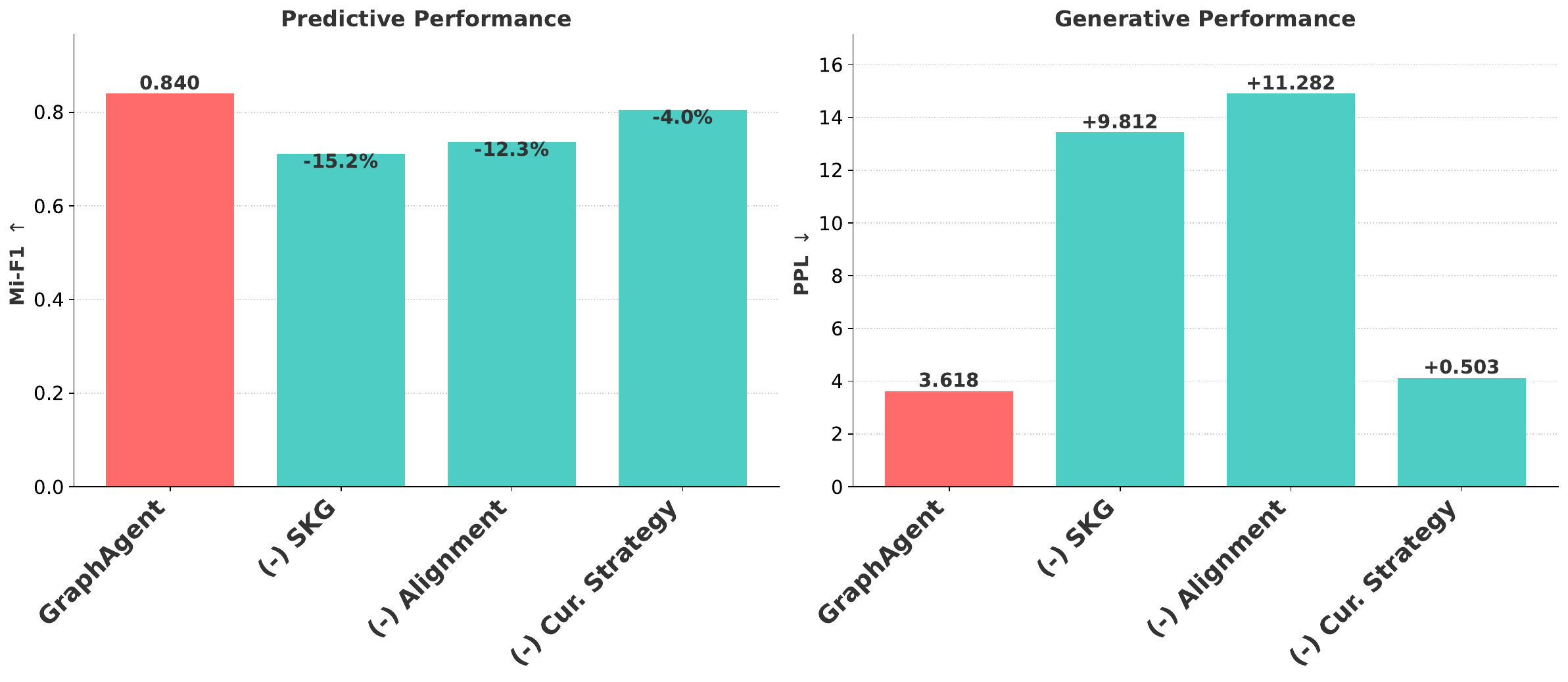}
    \vspace{-0.2in}
    \caption{Ablation study comparing \model with its variants on both graph-related prediction and graph-enhanced text generation tasks.}
    \vspace{-0.15in}
    \label{fig:ablation}
\end{wrapfigure}

\subsection{Ablation Study}
To evaluate each component in \model, we conducted an ablation study with the following variants: $\bullet$ \textbf{(-) SKG}: Removes the graph generation agent and excludes semantic knowledge graph tokens from LLM input. $\bullet$ \textbf{(-) Alignment}: Omits the graph-instruction alignment tuning described in Section~\ref{sec:alignment}, training directly with instruction input-output pairs. $\bullet$ \textbf{(-) Cur. Strategy}: Eliminates the curriculum learning strategy for agent task training (Section~\ref{sec:multi-task}), instead training all tasks simultaneously across all epochs. Figure~\ref{fig:ablation} presents the comparative results between \model and its variants on both predictive and generative tasks. Our analysis reveals two key findings:

$\bullet$ For predictive tasks, semantic knowledge graphs generated by the graph generation agent show the strongest impact, as their supplementary information substantially enhances model performance. In contrast, for generative tasks, the alignment component proves crucial for maintaining high performance, likely because these tasks demand sophisticated reasoning capabilities, making alignment tuning essential for developing deeper graph-instruction understanding.

$\bullet$ The curriculum training strategy shows consistent improvements across both task types. By enabling gradual progression from simpler predictive tasks to more complex generative ones, this approach allows the model to more effectively assimilate knowledge from various graph-instruction pairs, resulting in more robust overall performance.

%% file: relate.tex
\section{Related Work}
\label{sec:relate}
\vspace{-0.15in}

\noindent \textbf{Graph Representation Learning} enables analysis of complex relationships through specialized graph embedding techniques~\cite{chen2020simple,wu2020comprehensive}. Graph Neural Networks serve as its foundation, capturing node dependencies through message-passing mechanisms~\cite{dwivedi2023benchmarking,huang2024uncertainty}. Key architectures include Graph Convolutional Networks (GCNs)\cite{kipf2016semi,jin2020graph,wu2024graph}, which use localized convolutions for neighbor aggregation, and Graph Attention Networks (GAT)\cite{velivckovic2017graph,zhang2022graph,hao2023gat}, which incorporate attention mechanisms to weigh neighboring nodes' importance. In our \model, GNNs act as graph tokenizers, facilitating effective integration with LLMs. \\\vspace{-0.12in}

\noindent \textbf{Graph Language Models}. With the success of Large Language Models (LLMs), recent studies have focused on enhancing the generalization capabilities of graph models by integrating LLMs with Graph Neural Networks (GNNs)~\cite{tang2024higpt}. For instance, GraphGPT~\cite{tang2024graphgpt} enables LLMs to understand graph structural information by combining a graph encoder with an LLM through an alignment projector. LLaGA~\cite{chenllaga} enhances LLM capabilities for graph data by reorganizing nodes into structure-aware sequences. Additionally, ZeroG~\cite{li2024zerog} has been developed for zero-shot transfer learning in graph learning, leveraging language models to achieve effective cross-dataset generalization. However, most current graph language models primarily focus on capturing the topological information of explicit graph connections for standard representation learning tasks. In this work, we introduce a fully automated and easy-to-use agent framework that goes beyond traditional graph language models. Our framework is designed to tackle complex real-world data scenarios, which often involve both explicit relational graph connections and implicit graph-enhanced semantic dependencies. This allows us to address various downstream applications, including both graph-related predictive and text generative tasks. \\\vspace{-0.12in}

\noindent \textbf{LLM-empowered Agents}. LLM-empowered agents enhance user interactions by connecting complex data with intuitive communication. They utilize LLMs to efficiently integrate diverse information, allowing them to handle a broader range of tasks~\cite{shinn2024reflexion,xie2023openagents}. For example, language-based assistants use LLMs to combine reasoning with task-specific actions, improving performance in language understanding and decision-making~\cite{yaoreact,jimenezswe}. Vision-based assistants analyze visual data with LLMs to provide contextual insights, enhancing interactions with visual information~\cite{koh2024visualwebarena,hong2024cogagent}. Embodied agents particularly benefit from LLMs, gaining the ability to navigate complex environments and engage meaningfully with users, which enhances their application in robotics and smart systems~\cite{brehmer2024edgi,huang2024grounded}. However, a gap remains for agents that can understand relational data alongside rich textual information. This work aims to address that gap by developing an automated framework that integrates relational and textual data for various predictive and generative tasks.

%% file: conclusion.tex
\section{Conclusion}\vspace{-0.1in}
\label{sec:conclusion}

This work introduces a multi-agent framework that seamlessly integrates graph-based reasoning with advanced language modeling, effectively addressing complex language assistant scenarios involving both relational and textual data. The proposed \model\ features a dynamic pipeline that automates the understanding of graph-enhanced relational and textual semantics for both predictive and generative tasks. The framework consists of three key components: a graph generator agent that uncovers intricate semantic interdependencies, a task planning agent that interprets user queries, and a task execution agent that efficiently carries out tasks. This innovative agentic workflow enhances the adaptability of large language models to diverse datasets and significantly improves performance in benchmarking graph prediction tasks as well as in open-ended text generation tasks. In future work, we plan to extend our framework to incorporate visual information from multi-modal data, enabling it to better understand and generate content that integrates relational, textual and visual elements.

%% file: appendix.tex
\appendix \section{Appendix}
\label{sec:appendix}

\subsection{Detailed Implementation of \model}
\label{app:impl}
To ensure reproducibility of our experimental results, we provide comprehensive implementation details and technical specifications of our \model framework in this section.

\subsubsection{\bf System prompts of \model}
Tables~\ref{tab:sys_prompts} and~\ref{tab:llm_as_judge_prompt} present the comprehensive system prompts used in our framework. Specifically, Table~\ref{tab:sys_prompts} details the system prompts for the three core components: the \emph{Task Planning Agent}, the \emph{Graph Generation Agent}, and the task-specific prompt builders for the \emph{Multi-Task Agent}. Additionally, Table~\ref{tab:llm_as_judge_prompt} outlines the system prompts employed in our LLM-as-judge evaluation protocol.

\begin{table*}[htbp]
    \centering
    \small
    \caption{System Prompts for LLM-based Performance Evaluation.}
    \label{tab:llm_as_judge_prompt}
    \begin{tabular}{p{0.95\textwidth}}
    \toprule
    \midrule
    You are a professional researcher in computer science, AI. You are good at reading and judging written drafts of research papers. Now, your task is to judge between two paragraphs of "related work" section of the same topic. You have to decide which paragraph is written in a better way in terms of the following criteria: 1. It should strictly cover all the references provided. It is not acceptable if only part of the references is cited. 2. It is encouraged that the written content discusses how the related works differ from each other. 3. It is encouraged that detailed technical information is discussed for each reference. 4. It is encouraged that concise, neutral, and objective language is used. Here are the two paragraphs with the same topic <topic>: A:<content a>; B:<content b> Directly give your answers as ```A is better```, ```B is better``` or ```On par```(Use this very sparingly). Then, give a very short reasoning and reflection on why you think so.\\
    \bottomrule
    \end{tabular}
\end{table*}

\begin{wrapfigure}{R}{0.5\textwidth}
    \centering
    \small
    \vspace{-0.1in}
    \caption{Curriculum Learning Strategy: Training Epochs and Data Mixing Ratios.}
    \vspace{-0.1in}
    \label{tab:cur}
    \resizebox{\linewidth}{!}{
    \begin{tabular}{lccc}
        \toprule
        & {Alignment Data} & {Predictive Data} & {Generative Data} \\
        \midrule
        {Epoch 1} & 10\% & 70\% & 20\% \\
        {Epoch 2} & 5\% & 60\% & 35\% \\
        {Epoch 3} & 0\% & 50\% & 50\% \\
        {Afterwards} &  0\%& 40\% & 60\% \\
        \bottomrule
    \end{tabular}
    \vspace{-0.2in}
    }
\end{wrapfigure}

\subsubsection{\bf Curriculum training strategy}
We employ a curriculum learning strategy to effectively train our graph language model for multi-task scenarios. As shown in Table~\ref{tab:cur}, the training process begins with fundamental tasks to establish basic graph-language understanding, then progressively introduces more challenging components - advancing from predictive tasks to generative tasks. This graduated approach ensures robust model development and optimal performance across diverse task requirements.

\begin{table*}[htbp]
    \centering
    \small
    \vspace{-0.1in}
    \caption{Examples for the system prompts and system prompt builder we used in \model}
    \vspace{-0.1in}
    \label{tab:sys_prompts}
    \begin{tabular}{p{1.\textwidth}}
    \toprule
    \midrule
    \textbf{System prompt for intent and task parsing [$\mathbf{x}_\text{system\_tp}$]} \\
    You are very powerful assistant for graph-related tasks for diverse user inputs.
    You can do great in parsing the following important properties from the user input: 1. "graph source". This is either the uploaded file paths if the user uploads pre-defined graph for the task, or the user input contents as texts or documents that contain knowledge. 2. "graph task". the graph task type to handle, must be one of {"predictive\_predefined"}, {"predictive\_wild"}, {"open\_generation"}. You should infer the graph task to handle from the user input. 3. "user annotations". Any additional information the user provided in the query prompt. Could be task description,  label candidates or specific requirements. You are provided with two realistic examples to help you excel in the task: <few shot examples>.
\\
    \midrule
    \textbf{System prompt for scaffold knowledge node extraction at the 0-th step of the Graph Generation Agent [$\mathbf{x}_\text{system\_sk\_0}$]} \\
    You are very powerful assistant for graph-related tasks for diverse user inputs. You can do great in detecting and extracting the important scaffold nodes from the user input. A list of scaffold nodes reflect the top-level concepts or entities in the content, that are useful to form a knowledge graph for the content. You should carefully examine the input content to decide your extraction strategy. 1. For a general long document of a certain scenario, consider several most high-level aspects that are useful to grasp the key concepts in the document. Do not propose too specific concepts as scaffold nodes. It is very vital to be general and be abstract in your proposed scaffold nodes. 2. For inputs that are more formatted and contain specific entities, relationships, or concepts, you can directly adopt the key entities or concepts listed in the input as scaffold nodes. In this case, it is essential to accurately concentrate on the high-level formatted concepts or entities. For your output, use auto-increment ids to number the scaffold nodes, and infer the general type for each. You are provided with several examples to help you excel in the task: <few shot examples>. \\
    \midrule
    \textbf{System prompt for scaffold knowledge node extraction after the 0-th step of the Graph Generation Agent [$\mathbf{x}_\text{system\_sk\_1}$]} \\
    You are very powerful assistant for graph-related tasks for diverse user inputs. You can do great in detecting and extracting the important scaffold nodes from the user input. A list of scaffold nodes should be informative and representative of the key points in the text, that are useful to form a knowledge graph for the content. You should carefully examine the input content to decide your extraction strategy. You also need to provide a description of the extracted keywords for each scaffold node. The description should be detailed and informative, and can contain two parts: 1) a brief description of the keywords based on the contexts in the text, and 2) a detailed description of the keywords based on your own knowledge.
 You are provided with several examples to help you excel in the task: <few shot examples>. \\
\midrule
 \textbf{System prompt for knowledge description augmentation of the Graph Generation Agent [$\mathbf{x}_\text{system\_ka}$]} \\
You are a powerful assistant in generating information textual descriptions for a list of scaffold nodes. Each scaffold node represents a high-level key point or topic in the text, and your goal is to provide comprehensive and detailed texts related to each scaffold node. The texts can be from your own knowledge base with references to the original input content. Texts should be detailed and you should never miss any important information.
You can never miss any node in the input.
You should parse corresponding texts for each scaffold node in the input. You should always return the same number of scaffold nodes as the input.
 You are provided with several examples to help you excel in the task: <few shot examples>. \\
 \midrule
\textbf{System prompt builder template for graph multi task agent [$\mathbf{x}_\text{system\_ka}$]} \\
You are a powerful assistant in accomplishing diverse user required tasks with the help of structured knowledge as graphs. The current user requested task is of type: <$\mathbf{t}_\text{user}$>. The detailed request or provided information is: <$\mathbf{x}_\text{user\_ann}, \mathbf{g}_s$>. [\textbf{If predictive in the wild or open generation:}] For the required task, a heterogeneous knowledge graph is built to assist you as useful and informative knowledge references. There are <num. of meta types> types of nodes and edges in the graph, separately: <meta types>. The graph tokens for each type are: [<meta type>: <graph>]. [\textbf{If predictive with pre-defined graphs:}] For the required task, a pre-defined heterogeneous graph is provided as information reference. There are <num. of meta types> types of nodes and edges in the graph, separately: <meta types>. The graph tokens for each type are: [<meta type>: <graph>]. Additionally, a heterogeneous knowledge graph is also constructed to augment your knowledge for the task. There are <num. of meta types> types of nodes and edges in the graph, separately: <meta types>. The graph tokens for each type are: [<meta type>: <graph>]. Please generate response that satisfies the user's request.<$\mathbf{x}_\text{user\_ann}$>.  Provide concise reasoning if the task involves certain prediction. \\
    \bottomrule
    \end{tabular}
    \vspace{-0.1in}
\end{table*}

\begin{table*}[htbp]
    \centering
    \small
    \vspace{-0.1in}
    \caption{Examples for the system prompts and system prompt builder we used in \model}
    \vspace{-0.1in}
    \label{tab:sys_prompts}
    \begin{tabular}{p{1.\textwidth}}
    \toprule
    \midrule
    \textbf{System prompt for intent and task parsing [$\mathbf{x}_\text{system\_tp}$]} \\
    You are very powerful assistant for graph-related tasks for diverse user inputs.
    You can do great in parsing the following important properties from the user input: 1. "graph source". This is either the uploaded file paths if the user uploads pre-defined graph for the task, or the user input contents as texts or documents that contain knowledge. 2. "graph task". the graph task type to handle, must be one of {"predictive\_predefined"}, {"predictive\_wild"}, {"open\_generation"}. You should infer the graph task to handle from the user input. 3. "user annotations". Any additional information the user provided in the query prompt. Could be task description,  label candidates or specific requirements. You are provided with two realistic examples to help you excel in the task: <few shot examples>.
\\
    \midrule
    \textbf{System prompt for scaffold knowledge node extraction at the 0-th step of the Graph Generation Agent [$\mathbf{x}_\text{system\_sk\_0}$]} \\
    You are very powerful assistant for graph-related tasks for diverse user inputs. You can do great in detecting and extracting the important scaffold nodes from the user input. A list of scaffold nodes reflect the top-level concepts or entities in the content, that are useful to form a knowledge graph for the content. You should carefully examine the input content to decide your extraction strategy. 1. For a general long document of a certain scenario, consider several most high-level aspects that are useful to grasp the key concepts in the document. Do not propose too specific concepts as scaffold nodes. It is very vital to be general and be abstract in your proposed scaffold nodes. 2. For inputs that are more formatted and contain specific entities, relationships, or concepts, you can directly adopt the key entities or concepts listed in the input as scaffold nodes. In this case, it is essential to accurately concentrate on the high-level formatted concepts or entities. For your output, use auto-increment ids to number the scaffold nodes, and infer the general type for each. You are provided with several examples to help you excel in the task: <few shot examples>. \\
    \midrule
    \textbf{System prompt for scaffold knowledge node extraction after the 0-th step of the Graph Generation Agent [$\mathbf{x}_\text{system\_sk\_1}$]} \\
    You are very powerful assistant for graph-related tasks for diverse user inputs. You can do great in detecting and extracting the important scaffold nodes from the user input. A list of scaffold nodes should be informative and representative of the key points in the text, that are useful to form a knowledge graph for the content. You should carefully examine the input content to decide your extraction strategy. You also need to provide a description of the extracted keywords for each scaffold node. The description should be detailed and informative, and can contain two parts: 1) a brief description of the keywords based on the contexts in the text, and 2) a detailed description of the keywords based on your own knowledge.
 You are provided with several examples to help you excel in the task: <few shot examples>. \\
\midrule
 \textbf{System prompt for knowledge description augmentation of the Graph Generation Agent [$\mathbf{x}_\text{system\_ka}$]} \\
You are a powerful assistant in generating information textual descriptions for a list of scaffold nodes. Each scaffold node represents a high-level key point or topic in the text, and your goal is to provide comprehensive and detailed texts related to each scaffold node. The texts can be from your own knowledge base with references to the original input content. Texts should be detailed and you should never miss any important information.
You can never miss any node in the input.
You should parse corresponding texts for each scaffold node in the input. You should always return the same number of scaffold nodes as the input.
 You are provided with several examples to help you excel in the task: <few shot examples>. \\
 \midrule
\textbf{System prompt builder template for graph multi task agent [$\mathbf{x}_\text{system\_ka}$]} \\
You are a powerful assistant in accomplishing diverse user required tasks with the help of structured knowledge as graphs. The current user requested task is of type: <$\mathbf{t}_\text{user}$>. The detailed request or provided information is: <$\mathbf{x}_\text{user\_ann}, \mathbf{g}_s$>. [\textbf{If predictive in the wild or open generation:}] For the required task, a heterogeneous knowledge graph is built to assist you as useful and informative knowledge references. There are <num. of meta types> types of nodes and edges in the graph, separately: <meta types>. The graph tokens for each type are: [<meta type>: <graph>]. [\textbf{If predictive with pre-defined graphs:}] For the required task, a pre-defined heterogeneous graph is provided as information reference. There are <num. of meta types> types of nodes and edges in the graph, separately: <meta types>. The graph tokens for each type are: [<meta type>: <graph>]. Additionally, a heterogeneous knowledge graph is also constructed to augment your knowledge for the task. There are <num. of meta types> types of nodes and edges in the graph, separately: <meta types>. The graph tokens for each type are: [<meta type>: <graph>]. Please generate response that satisfies the user's request.<$\mathbf{x}_\text{user\_ann}$>.  Provide concise reasoning if the task involves certain prediction. \\
    \bottomrule
    \end{tabular}
    \vspace{-0.1in}
\end{table*}

\begin{table*}[htbp]
    \centering
    \small
    \caption{Qualitative comparison for a GovReport task between \model, llama3-8b and Llama3-70b.}
    \vspace{-0.1in}
    \label{tab:govreport_exp_1}
    \begin{tabular}{p{0.95\textwidth}}
    \toprule
    \midrule
    \textbf{[User]}\\
    Here is a government report concerning specific topics. Your task is to write a paragraph that highly condenses the content without losing any important information. \newline
    Consistent with the discretion afforded by the APA, Regulations.gov and agency-specific comment websites use required and optional fields on comment forms to collect some identity information from commenters. In addition to the text of the comment, agencies may choose to collect identity information by requiring commenters to fill in other fields, such as name, address, and email address before they are able to submit a comment. Regardless of the fields required by the comment form, the selected agencies all accept anonymous comments in practice. Specifically, in the comment forms on Regulations.gov and agency- specific comment websites, a commenter can submit under a fictitious name, such as \"Anonymous Anonymous,\" enter a single letter in each required field, or provide a fabricated address. In each of these scenarios, as long as a character or characters are entered into the required fields, the comment will be accepted. Further, because the APA does not require agencies to authenticate submitted identity information, neither Regulations.gov nor the agency-specific comment websites contain mechanisms to check the validity of identity information that commenters submit through comment forms. Regulations.gov and agency-specific comment websites also collect some information about public users' interaction with their websites through application event logs and proxy server logs, though the APA does not require agencies to collect or verify it as part of the rulemaking process. This information, which can include a public user's Internet Protocol (IP) address, browser type and operating system, and the time and date of webpage visits, is collected separately from the comment submission process as part of routine information technology management for system security and performance, and cannot be reliably connected to specific comments. Seven of the 10 selected agencies have documented some internal guidance associated with the identity of commenters during the three phases of the public comment process: intake, analysis, and response to comments. However, the focus and substance of this guidance varies by agency and phase of the comment process. As shown in Table 1, for selected agencies that have guidance associated with the identity of commenters, it most frequently relates to the comment intake or response to comment phases of the public comment process. The guidance for these phases addresses activities such as managing duplicate comments (those with identical or near-identical comment text but varied identity information) or referring to commenters in a final rule...\\
    \midrule
    \midrule
    \textbf{[GraphAgent]}\\
    The report \colorbox{green!30}{examines} how government agencies manage and disclose identity information associated with public comments on proposed regulations. \colorbox{green!30}{It finds} that while agencies are not required to collect or verify identity information, they do collect some information from commenters and accept anonymous comments. \colorbox{green!30}{However, the report notes} inconsistencies in how agencies post identity information, particularly with duplicate comments. \colorbox{green!30}{It also highlights} a lack of transparency in disclosing these practices to the public, potentially hindering meaningful participation in the rulemaking process. \colorbox{green!30}{The report recommends} that agencies establish clear policies for posting comments and communicate these practices to the public more effectively. \colorbox{green!30}{Since issuing the report}, some agencies have begun implementing these recommendations, but others have not yet done so.\\
    \midrule
    \textbf{[Llama3-8b]}\\
    The Government Accountability Office (GAO) report examines the treatment of identity information by 10 federal agencies during the public comment process. While agencies are not required to collect or verify identity information, they may choose to do so through comment forms. However, the report finds that agencies vary in their treatment of identity information, with some accepting anonymous comments and others requiring commenters to provide name, address, and email address. The report also notes that agencies do not consistently post duplicate comments, which can lead to inconsistent presentation of identity information to the public. Additionally, agencies do not provide clear information about their posting practices, which can limit public users' ability to explore and use the data. The GAO recommends that agencies establish policies for posting comments and provide clear communication to the public about their posting practices, particularly with regard to identity information and duplicate comments. Since the report's issuance, some agencies have taken steps to address these recommendations.\\
    \midrule
    \textbf{[Llama3-70b]}\\
    The government's public comment process, facilitated by Regulations.gov and agency-specific websites, collects identity information from commenters, but allows anonymous comments and does not verify submitted information. While agencies may require certain fields to be filled, commenters can submit fictitious information or minimal data, and comments will still be accepted. The treatment of identity information varies across agencies, with some posting all comments and others posting only a sample of duplicate comments, which can lead to inconsistent presentation of identity information to the public. Furthermore, agencies do not effectively communicate their posting practices and limitations to the public, potentially leading to inaccurate conclusions about the comments submitted. To address this, recommendations have been made to establish policies for posting comments and clearly communicate them to the public, with some agencies having taken action to implement these changes.\\
    \bottomrule
    \end{tabular}
\end{table*}

\begin{table*}[htbp]
    \centering
    \small
    \caption{Qualitative comparison for a NLP-Related Works task between \model, fine-tuned llama3-8b and Llama3-70b.}
    \vspace{-0.1in}
    \label{tab:nlp_rw_exp_1}
    \begin{tabular}{p{0.95\textwidth}}
    \toprule
    \midrule
    \textbf{[User]}\\
    I am writing a "related works" section for my research paper in the NLP domain. The topic to cover is "Dense Passage Retrieval". I would like to discuss the following papers in the section:\newline
    Title: Representation Degeneration Problem in Training Natural Language Generation Models. Abstract: ... \newline
    Title: Deeper Text Understanding for IR with Contextual Neural Language Modeling. Abstract: ... \newline
    Title: Approximate Nearest Neighbor Negative Contrastive Learning for Dense Text Retrieval. Abstract: ... \newline
    Title: Unsupervised Corpus Aware Language Model Pre-training for Dense Passage Retrieval. Abstract: ... \newline
    Title: PAIR: Leveraging Passage-Centric Similarity Relation for Improving Dense Passage Retrieval. Abstract:... \newline
    Further, 
    Use @CITE[id]@ to cite a paper in the paragraph. For example, @CITE[1]@ will be replaced by the citation of the paper with id 1. At the end of the generated content, list all the references in the format of ID: Title.
    \\
    \midrule
    \midrule
    \textbf{[\model]}\\
Dense passage retrieval (DPR) has recently attracted a lot of attention in the IR community, which aims to retrieve relevant passages from a large corpus with the help of a query. DPR is based on the neural network, which can learn query and passage representations from the corpus and query separately. The retrieval model is usually trained by contrastive loss, which is to pull the query representation close to the passage representation when they are relevant, and push them apart when they are irrelevant. \colorbox{green!30}{In the early stage}, DPR is mainly based on the BERT model @CITE[4]@, which is pre-trained by masked language modeling. \colorbox{green!30}{However, BERT is} prone to degenerate into a linear mapping to the word embeddings @CITE[5]@, which leads to a performance drop. \colorbox{green!30}{To alleviate this problem}, several methods are proposed to train DPR model. \colorbox{green!30}{For example}, @CITE[3]@ propose the use of the negative sampling strategy, which is to sample a negative passage from the corpus to calculate the contrastive loss. @CITE[1]@ propose the use of the nearest neighbor negative sampling strategy, which is to sample a negative passage from the corpus with the nearest distance to the query. @CITE[2]@ propose the use of the contrastive loss with the corpus-level information, which is to pull the passage representation close to the corpus-level representation when they are relevant.\newline
References: ID: 1. Title: Representation Degeneration Problem in Training Natural Language Generation Models\newline
ID: 2. Title: Deeper Text Understanding for IR with Contextual Neural Language Modeling\newline
ID: 3. Title: Approximate Nearest Neighbor Negative Contrastive Learning for Dense Text Retrieval\newline
ID: 4. Title: Unsupervised Corpus Aware Language Model Pre-training for Dense Passage Retrieval\newline
ID: 5. Title: PAIR: Leveraging Passage-Centric Similarity Relation for Improving Dense Passage Retrieval
    \\
    \midrule
    \textbf{[Llama3-70b]} \\
    Dense passage retrieval has gained significant attention in recent years, leveraging the power of neural networks to learn complex language patterns and query-document relations @CITE[2]@. This approach has many intriguing advantages over sparse retrieval, such as better capturing of language structures and improved performance on queries written in natural languages @CITE[2]@. However, the effectiveness of dense retrieval often requires combination with sparse retrieval, and the training mechanisms used in dense retrieval can be a bottleneck @CITE[3]@. To address this, various techniques have been proposed, including Approximate Nearest Neighbor Negative Contrastive Learning @CITE[3]@, unsupervised corpus-aware language model pre-training @CITE[4]@, and leveraging passage-centric similarity relations @CITE[5]@. These advancements have significantly improved the performance of dense passage retrieval models, making them a promising approach for various natural language processing tasks.\\
    \midrule
    \textbf{[Llama3-8b Fine-tuned]} \\
    Recent advancements in PLMs such as BERT \colorbox{red!30}{(De- vlin et al., 2018), Roberta (@CITE[1]@, and GPT (@CITE[3]@} have enabled numerous unsupervised techniques to derive dense represen- tations of queries and passages for retrieval. These approaches primarily use a Bi-Encoder structure to embed text in a low-dimensional space and learn similarity relations via contrastive learning, con- trasting traditional sparse retrieval methods like BM25 or DeepCT (@CITE[2]@;. \colorbox{red!30}{DPR ( pi- oneered an unsupervised} dense passage retrieval framework, affirming the feasibility of using dense representations for retrieval independently. This efficient and operational approach was further refined by subsequent studies \colorbox{red!30}{(@CITE[4]@; ; ; @CITE[5]@} that focused on high-quality negative sample min- ing, additional passage relation analysis, and extra training. The essence of these methods is to repre- sent texts in a target space where queries are closer to relevant and distant from irrelevant passages.\newline
References: ID: 1. Title: Representation Degeneration Problem in Training Natural Language Generation Models
ID: 2. Title: Deeper Text Understanding for IR with Contextual Neural Language Modeling
ID: 3. Title: Approximate Nearest Neighbor Negative Contrastive Learning for Dense Text Retrieval
ID: 4. Title: Unsupervised Corpus Aware Language Model Pre-training for Dense Passage Retrieval
ID: 5. Title: PAIR: Leveraging Passage-Centric Similarity Relation for Improving Dense Passage Retrieval\\
    \bottomrule
    \end{tabular}
\end{table*}

\begin{table*}[htbp]
    \centering
    \small
    \caption{Visualization examples for automatically discovered semantic knowledge graphs for a GovReport sample.}
    \vspace{-0.1in}
    \label{tab:skg_example_1}
    \begin{tabular}{p{0.95\textwidth}}
    \toprule
    \midrule
    \textbf{[Original Text with Knowledge]}\\
    In our June 2019 report, we found that, while abuse deficiencies cited in nursing homes were relatively rare from 2013 through 2017, they became more frequent during that time, with the largest increase in severe cases. Specifically, abuse deficiencies comprised less than 1 percent of the total deficiencies in each of the years we examined, which is likely conservative. Abuse in nursing homes is often underreported by residents, family, staff, and the state survey agency, according to CMS officials and stakeholders we interviewed. However, abuse deficiencies more than doubled--from 430 in 2013 to 875 in 2017--over the 5-year period. (See appendix II.) In addition, abuse deficiencies cited in 2017 were more likely to be categorized at the highest levels of severity-- deficiencies causing actual harm to residents or putting residents in immediate jeopardy--than they were in 2013. In light of the increased number and severity of abuse deficiencies, it is imperative that CMS have strong nursing home oversight in place to protect residents from abuse; however, we found oversight gaps that may limit the agency's ability to do so. Specifically, we found that CMS: (1) cannot readily access data on the type of abuse or type of perpetrator, (2) has not provided guidance on what information nursing homes should include in facility-reported incidents, and (3) has numerous gaps in its referral process that can result in delayed and missed referrals to law enforcement. In our June 2019 report, we found that CMS's data do not allow for the type of abuse or perpetrator to be readily identified by the agency. Specifically, CMS does not require the state survey agencies to record abuse and perpetrator type and, when this information is recorded, it cannot be easily analyzed by CMS. Therefore, we reviewed a representative sample of 400 CMS narrative descriptions--written by state surveyors--associated with abuse deficiencies cited in 2016 and 2017 to identify the most common types of abuse and perpetrators. From this review, we found that physical abuse (46 percent) and mental/verbal abuse (44 percent) occurred most often in nursing homes, followed by sexual abuse (18 percent). Furthermore, staff, which includes those working in any part of the nursing home, were more often the perpetrators (58 percent) of abuse in deficiency narratives, followed by resident perpetrators (30 percent) and other types of perpetrators (2 percent). (See appendix III for examples from our abuse deficiency narrative review.)...\\
    \midrule
    \textbf{[Scaffold Node $k=0$] (top-level aspects or concepts)}\\
    Node meta type: \colorbox{green!30}{Policy Objectives And Goals} \\
    Text attribute: The policy objectives and goals of this report are to ensure that CMS has strong nursing home oversight in place to protect residents from abuse. The report aims to identify and address the gaps in CMS's ability to monitor and respond to abuse deficiencies effectively.\\
    Node meta type: \colorbox{green!30}{Stakeholder Impact and Implications} \\
    Text attribute: Stakeholders impacted by this report include nursing home residents, their families, staff, state survey agencies, and CMS officials. The implications are that without proper oversight and reporting mechanisms, abuse in nursing homes may continue to be underreported and inadequately addressed.
    Node meta type: \colorbox{green!30}{Methodology and Evidence} \\
    Text attribute: The methodology involved reviewing a representative sample of 400 CMS narrative descriptions associated with abuse deficiencies cited in 2016 and 2017. This review aimed to identify the most common types of abuse and perpetrators. Additionally, interviews with CMS officials and stakeholders provided insights into the underreporting and handling of abuse incidents.\\
    Node meta type: \colorbox{green!30}{Findings and Recommendations}\\
    Text attribute: Key findings include the underreporting of abuse in nursing homes, the doubling of abuse deficiencies from 2013 to 2017, and the increased severity of these deficiencies. Recommendations include requiring state survey agencies to report abuse and perpetrator type, providing guidance on facility-reported incidents, and improving the referral process to law enforcement.\\
    Node meta type: \colorbox{green!30}{Implementation and Evaluation}\\
    Text attribute: As of November 2019, the Department of Health and Human Services (HHS) had not implemented the recommendations made in the report. The implementation and evaluation of these recommendations are crucial to improving CMS's oversight and protecting nursing home residents from abuse.\\
    \midrule
    \textbf{[Scaffold Node $k=1$] (Fine-grained concepts as keywords. Text attributes are omitted.)} \\
    \colorbox{blue!30}{abuse deficiencies in nursing homes}; \colorbox{blue!30}{Centers for Medicare \& Medicaid Services (CMS)}; \colorbox{blue!30}{oversight gaps}; \colorbox{blue!30}{recommendations for improvement}; \colorbox{blue!30}{nursing home oversight}; \colorbox{blue!30}{resident protection}; \colorbox{blue!30}{abuse deficiencies}; \colorbox{blue!30}{CMS's ability to monitor}; \colorbox{blue!30}{nursing home residents}; \colorbox{blue!30}{abuse in nursing homes}; \colorbox{blue!30}{state survey agencies}; \colorbox{blue!30}{CMS officials}; \colorbox{blue!30}{CMS narrative descriptions}; \colorbox{blue!30}{abuse deficiencies}; \colorbox{blue!30}{underreporting of abuse incidents}; \colorbox{blue!30}{CMS officials and stakeholders}; \colorbox{blue!30}{underreporting of abuse}; \colorbox{blue!30}{abuse deficiencies}; \colorbox{blue!30}{state survey agencies}; \colorbox{blue!30}{referral process to law enforcement}; \colorbox{blue!30}{Department of Health and Human Services}; \colorbox{blue!30}{Centers for Medicare \& Medicaid Services}; \colorbox{blue!30}{nursing home residents}; \colorbox{blue!30}{abuse}; 
    \\
    \bottomrule
    \end{tabular}
\end{table*}

\begin{table*}[htbp]
    \centering
    \small
    \caption{Visualization examples for automatically discovered semantic knowledge graphs for an Arxiv sample, similarly for NLP-Related Works data.}
    \vspace{-0.1in}
    \label{tab:skg_example_2}
    \begin{tabular}{p{0.95\textwidth}}
    \toprule
    \midrule
    \textbf{[Original Text with Knowledge]}\\
    Title: A Simple Zero-shot Prompt Weighting Technique to Improve Prompt Ensembling in Text-Image Models. Abstract: Contrastively trained text-image models have the remarkable ability to perform zero-shot classification, that is, classifying previously unseen images into categories that the model has never been explicitly trained to identify. However, these zero-shot classifiers need prompt engineering to achieve high accuracy. Prompt engineering typically requires hand-crafting a set of prompts for individual downstream tasks. In this work, we aim to automate this prompt engineering and improve zero-shot accuracy through prompt ensembling. In particular, we ask "Given a large pool of prompts, can we automatically score the prompts and ensemble those that are most suitable for a particular downstream dataset, without needing access to labeled validation data?". We demonstrate that this is possible. In doing so, we identify several pathologies in a naive prompt scoring method where the score can be easily overconfident due to biases in pre-training and test data, and we propose a novel prompt scoring method that corrects for the biases. Using our proposed scoring method to create a weighted average prompt ensemble, our method outperforms equal average ensemble, as well as hand-crafted prompts, on ImageNet, 4 of its variants, and 11 fine-grained classification benchmarks, all while being fully automatic, optimization-free, and not requiring access to labeled validation data.\\
    \midrule
    \textbf{[Scaffold Node $k=0$] (top-level aspects or concepts)}\\
    Node meta type: \colorbox{green!30}{Research Background} \\
    Text attribute: Contrastively trained text-image models possess the ability to perform zero-shot classification, which involves categorizing unseen images into untrained categories. However, achieving high accuracy in zero-shot classification often requires meticulous prompt engineering, typically involving hand-crafted prompts tailored for specific downstream tasks.\\
    Node meta type: \colorbox{green!30}{Research Question} \\
    Text attribute: The research question addressed in this work is whether it is possible to automatically score and ensemble the most suitable prompts from a large pool for a particular downstream dataset, without relying on labeled validation data. This question stems from the need to automate and improve the accuracy of zero-shot classification through better prompt engineering.\\
    Node meta type: \colorbox{green!30}{Methodology} \\
    Text attribute: The methodology involves identifying and addressing pathologies in a naive prompt scoring method, which can be overly confident due to biases in pre-training and test data. The authors propose a novel prompt scoring method that corrects for these biases, enabling the creation of a weighted average prompt ensemble that is fully automatic and optimization-free.\\
    Node meta type: \colorbox{green!30}{Key Results} \\
    Text attribute: The key results demonstrate that the proposed prompt weighting technique outperforms both equal average ensemble and hand-crafted prompts on ImageNet, four of its variants, and 11 fine-grained classification benchmarks. The method achieves this while remaining fully automatic, not requiring optimization, and without access to labeled validation data.\\
    \textbf{[Scaffold Node $k=1$] (Fine-grained concepts as keywords. Text attributes are omitted.)} \\
    \colorbox{blue!30}{zero-shot prompt weighting}; \colorbox{blue!30}{automating prompt engineering}; \colorbox{blue!30}{zero-shot classification accuracy}; \colorbox{blue!30}{zero-shot classification}; \colorbox{blue!30}{meticulous prompt engineering}; \colorbox{blue!30}{automatic scoring}; \colorbox{blue!30}{ensemble prompts}; \colorbox{blue!30}{zero-shot classification}; \colorbox{blue!30}{prompt engineering}; \colorbox{blue!30}{naive prompt scoring method}; \colorbox{blue!30}{novel prompt scoring method}; \colorbox{blue!30}{weighted average prompt ensemble}; \colorbox{blue!30}{prompt weighting technique}; \colorbox{blue!30}{fully automatic};
    \\
    \bottomrule
    \end{tabular}
\end{table*}

%% file: main.bbl
\begin{thebibliography}{54}
\providecommand{\natexlab}[1]{#1}
\providecommand{\url}[1]{\texttt{#1}}
\expandafter\ifx\csname urlstyle\endcsname\relax
  \providecommand{\doi}[1]{doi: #1}\else
  \providecommand{\doi}{doi: \begingroup \urlstyle{rm}\Url}\fi

\bibitem[Bengio et~al.(2009)Bengio, Louradour, Collobert, and Weston]{bengio2009curriculum}
Yoshua Bengio, J{\'e}r{\^o}me Louradour, Ronan Collobert, and Jason Weston.
\newblock Curriculum learning.
\newblock In \emph{ICML}, pp.\  41--48, 2009.

\bibitem[Brehmer et~al.(2024)Brehmer, Bose, De~Haan, and Cohen]{brehmer2024edgi}
Johann Brehmer, Joey Bose, Pim De~Haan, and Taco~S Cohen.
\newblock Edgi: Equivariant diffusion for planning with embodied agents.
\newblock \emph{NeurIPS}, 36, 2024.

\bibitem[Chen et~al.(2023)Chen, Zhang, Zhang, Han, Cheng, Li, Dong, and Tang]{chen2023web}
Bo~Chen, Jing Zhang, Fanjin Zhang, Tianyi Han, Yuqing Cheng, Xiaoyan Li, Yuxiao Dong, and Jie Tang.
\newblock Web-scale academic name disambiguation: the whoiswho benchmark, leaderboard, and toolkit.
\newblock In \emph{KDD}, pp.\  3817--3828, 2023.

\bibitem[Chen et~al.(2020)Chen, Wei, Huang, Ding, and Li]{chen2020simple}
Ming Chen, Zhewei Wei, Zengfeng Huang, Bolin Ding, and Yaliang Li.
\newblock Simple and deep graph convolutional networks.
\newblock In \emph{ICML}, pp.\  1725--1735. PMLR, 2020.

\bibitem[Chen et~al.(2024{\natexlab{a}})Chen, Zhao, Jaiswal, Shah, and Wang]{chen2024llaga}
Runjin Chen, Tong Zhao, Ajay Jaiswal, Neil Shah, and Zhangyang Wang.
\newblock Llaga: Large language and graph assistant.
\newblock \emph{ICML}, 2024{\natexlab{a}}.

\bibitem[Chen et~al.(2024{\natexlab{b}})Chen, Zhao, JAISWAL, Shah, and Wang]{chenllaga}
Runjin Chen, Tong Zhao, AJAY~KUMAR JAISWAL, Neil Shah, and Zhangyang Wang.
\newblock Llaga: Large language and graph assistant.
\newblock In \emph{ICML}, 2024{\natexlab{b}}.

\bibitem[Dai et~al.(2022)Dai, Jin, Liu, and Wang]{dai2022towards}
Enyan Dai, Wei Jin, Hui Liu, and Suhang Wang.
\newblock Towards robust graph neural networks for noisy graphs with sparse labels.
\newblock In \emph{WSDM}, pp.\  181--191, 2022.

\bibitem[Dwivedi et~al.(2023)Dwivedi, Joshi, Luu, Laurent, Bengio, and Bresson]{dwivedi2023benchmarking}
Vijay~Prakash Dwivedi, Chaitanya~K Joshi, Anh~Tuan Luu, Thomas Laurent, Yoshua Bengio, and Xavier Bresson.
\newblock Benchmarking graph neural networks.
\newblock \emph{JMLR}, 24\penalty0 (43):\penalty0 1--48, 2023.

\bibitem[Fey \& Lenssen(2019)Fey and Lenssen]{fey2019fast}
Matthias Fey and Jan~Eric Lenssen.
\newblock Fast graph representation learning with pytorch geometric.
\newblock \emph{arXiv preprint arXiv:1903.02428}, 2019.

\bibitem[Fey et~al.(2023)Fey, Hu, Huang, Lenssen, Ranjan, Robinson, Ying, You, and Leskovec]{fey2023relational}
Matthias Fey, Weihua Hu, Kexin Huang, Jan~Eric Lenssen, Rishabh Ranjan, Joshua Robinson, Rex Ying, Jiaxuan You, and Jure Leskovec.
\newblock Relational deep learning: Graph representation learning on relational databases.
\newblock \emph{arXiv preprint arXiv:2312.04615}, 2023.

\bibitem[Fu et~al.(2020)Fu, Zhang, Meng, and King]{fu2020magnn}
Xinyu Fu, Jiani Zhang, Ziqiao Meng, and Irwin King.
\newblock Magnn: Metapath aggregated graph neural network for heterogeneous graph embedding.
\newblock In \emph{Proceedings of the web conference 2020}, pp.\  2331--2341, 2020.

\bibitem[Hamilton(2020)]{hamilton2020graph}
William~L Hamilton.
\newblock \emph{Graph representation learning}.
\newblock Morgan \& Claypool Publishers, 2020.

\bibitem[Hamilton et~al.(2017)Hamilton, Ying, and Leskovec]{graphsage}
William~L. Hamilton, Zhitao Ying, and Jure Leskovec.
\newblock Inductive representation learning on large graphs.
\newblock In \emph{{NeurIPS}}, pp.\  1024--1034, 2017.

\bibitem[Hao et~al.(2023)Hao, Huang, Feng, Yuan, and Li]{hao2023gat}
Qianyue Hao, Wenzhen Huang, Tao Feng, Jian Yuan, and Yong Li.
\newblock Gat-mf: Graph attention mean field for very large scale multi-agent reinforcement learning.
\newblock In \emph{KDD}, pp.\  685--697, 2023.

\bibitem[He et~al.(2023)He, Bresson, Laurent, Perold, LeCun, and Hooi]{he2023harnessing}
Xiaoxin He, Xavier Bresson, Thomas Laurent, Adam Perold, Yann LeCun, and Bryan Hooi.
\newblock Harnessing explanations: Llm-to-lm interpreter for enhanced text-attributed graph representation learning.
\newblock \emph{arXiv preprint arXiv:2305.19523}, 2023.

\bibitem[Hong et~al.(2024)Hong, Wang, Lv, Xu, Yu, Ji, Wang, Wang, Dong, Ding, et~al.]{hong2024cogagent}
Wenyi Hong, Weihan Wang, Qingsong Lv, Jiazheng Xu, Wenmeng Yu, Junhui Ji, Yan Wang, Zihan Wang, Yuxiao Dong, Ming Ding, et~al.
\newblock Cogagent: A visual language model for gui agents.
\newblock In \emph{CVPR}, pp.\  14281--14290, 2024.

\bibitem[Hu et~al.(2020)Hu, Dong, Wang, and Sun]{HGT}
Ziniu Hu, Yuxiao Dong, Kuansan Wang, and Yizhou Sun.
\newblock Heterogeneous graph transformer.
\newblock In \emph{{WWW}}, pp.\  2704--2710. {ACM} / {IW3C2}, 2020.

\bibitem[Huang et~al.(2024{\natexlab{a}})Huang, Jin, Candes, and Leskovec]{huang2024uncertainty}
Kexin Huang, Ying Jin, Emmanuel Candes, and Jure Leskovec.
\newblock Uncertainty quantification over graph with conformalized graph neural networks.
\newblock \emph{NeurIPS}, 36, 2024{\natexlab{a}}.

\bibitem[Huang et~al.(2024{\natexlab{b}})Huang, Xia, Shah, Driess, Zeng, Lu, Florence, Mordatch, Levine, Hausman, et~al.]{huang2024grounded}
Wenlong Huang, Fei Xia, Dhruv Shah, Danny Driess, Andy Zeng, Yao Lu, Pete Florence, Igor Mordatch, Sergey Levine, Karol Hausman, et~al.
\newblock Grounded decoding: Guiding text generation with grounded models for embodied agents.
\newblock \emph{NeurIPS}, 36, 2024{\natexlab{b}}.

\bibitem[Jimenez et~al.(2024)Jimenez, Yang, Wettig, Yao, Pei, Press, and Narasimhan]{jimenezswe}
Carlos~E Jimenez, John Yang, Alexander Wettig, Shunyu Yao, Kexin Pei, Ofir Press, and Karthik~R Narasimhan.
\newblock Swe-bench: Can language models resolve real-world github issues?
\newblock In \emph{ICLR}, 2024.

\bibitem[Jin et~al.(2020)Jin, Ma, Liu, Tang, Wang, and Tang]{jin2020graph}
Wei Jin, Yao Ma, Xiaorui Liu, Xianfeng Tang, Suhang Wang, and Jiliang Tang.
\newblock Graph structure learning for robust graph neural networks.
\newblock In \emph{KDD}, pp.\  66--74, 2020.

\bibitem[Kipf \& Welling(2017)Kipf and Welling]{kipf2016semi}
Thomas~N Kipf and Max Welling.
\newblock Semi-supervised classification with graph convolutional networks.
\newblock In \emph{ICLR}, 2017.

\bibitem[Koh et~al.(2024)Koh, Lo, Jang, Duvvur, Lim, Huang, Neubig, Zhou, Salakhutdinov, and Fried]{koh2024visualwebarena}
Jing~Yu Koh, Robert Lo, Lawrence Jang, Vikram Duvvur, Ming~Chong Lim, Po-Yu Huang, Graham Neubig, Shuyan Zhou, Ruslan Salakhutdinov, and Daniel Fried.
\newblock Visualwebarena: Evaluating multimodal agents on realistic visual web tasks.
\newblock \emph{ACL}, 2024.

\bibitem[Li et~al.(2023)Li, Wang, Li, Fu, Shen, Shang, and McAuley]{li2023text}
Jiacheng Li, Ming Wang, Jin Li, Jinmiao Fu, Xin Shen, Jingbo Shang, and Julian McAuley.
\newblock Text is all you need: Learning language representations for sequential recommendation.
\newblock In \emph{KDD}, pp.\  1258--1267, 2023.

\bibitem[Li et~al.(2024)Li, Wang, Li, Yu, and Li]{li2024zerog}
Yuhan Li, Peisong Wang, Zhixun Li, Jeffrey~Xu Yu, and Jia Li.
\newblock Zerog: Investigating cross-dataset zero-shot transferability in graphs.
\newblock In \emph{KDD}, pp.\  1725--1735, 2024.

\bibitem[Lin(2004)]{lin2004rouge}
Chin-Yew Lin.
\newblock Rouge: A package for automatic evaluation of summaries.
\newblock In \emph{Text summarization branches out}, pp.\  74--81, 2004.

\bibitem[Liu et~al.(2024)Liu, Li, Wu, and Lee]{liu2024visual}
Haotian Liu, Chunyuan Li, Qingyang Wu, and Yong~Jae Lee.
\newblock Visual instruction tuning.
\newblock \emph{NeurIPS}, 36, 2024.

\bibitem[Liu et~al.(2022)Liu, Wang, Bo, Shi, and Pei]{liu2022revisiting}
Nian Liu, Xiao Wang, Deyu Bo, Chuan Shi, and Jian Pei.
\newblock Revisiting graph contrastive learning from the perspective of graph spectrum.
\newblock \emph{NeurIPS}, 35:\penalty0 2972--2983, 2022.

\bibitem[Llama~Team(2024)]{dubey2024llama3herdmodels}
AI~@~Meta Llama~Team.
\newblock The llama 3 herd of models, 2024.
\newblock URL \url{https://arxiv.org/abs/2407.21783}.

\bibitem[Lu et~al.(2024)Lu, Yang, Li, Wang, and Wang]{lu2024llmscore}
Yujie Lu, Xianjun Yang, Xiujun Li, Xin~Eric Wang, and William~Yang Wang.
\newblock Llmscore: Unveiling the power of large language models in text-to-image synthesis evaluation.
\newblock \emph{NeurIPS}, 36, 2024.

\bibitem[Mao et~al.(2024)Mao, Chen, Tang, Zhao, Ma, Zhao, Shah, Galkin, and Tang]{mao2024graph}
Haitao Mao, Zhikai Chen, Wenzhuo Tang, Jianan Zhao, Yao Ma, Tong Zhao, Neil Shah, Michael Galkin, and Jiliang Tang.
\newblock Graph foundation models.
\newblock \emph{ICML}, 2024.

\bibitem[Papineni et~al.(2002)Papineni, Roukos, Ward, and Zhu]{papineni2002bleu}
Kishore Papineni, Salim Roukos, Todd Ward, and Wei-Jing Zhu.
\newblock Bleu: a method for automatic evaluation of machine translation.
\newblock In \emph{ACL}, pp.\  311--318, 2002.

\bibitem[Shinn et~al.(2023)Shinn, Cassano, Gopinath, Narasimhan, and Yao]{shinn2024reflexion}
Noah Shinn, Federico Cassano, Ashwin Gopinath, Karthik Narasimhan, and Shunyu Yao.
\newblock Reflexion: Language agents with verbal reinforcement learning.
\newblock \emph{NeurIPS}, 36, 2023.

\bibitem[Shuai et~al.(2022)Shuai, Zhang, Wu, Sun, Hong, Wang, and Li]{shuai2022review}
Jie Shuai, Kun Zhang, Le~Wu, Peijie Sun, Richang Hong, Meng Wang, and Yong Li.
\newblock A review-aware graph contrastive learning framework for recommendation.
\newblock In \emph{SIGIR}, pp.\  1283--1293, 2022.

\bibitem[Tang et~al.(2024{\natexlab{a}})Tang, Yang, Wei, Shi, Su, Cheng, Yin, and Huang]{tang2024graphgpt}
Jiabin Tang, Yuhao Yang, Wei Wei, Lei Shi, Lixin Su, Suqi Cheng, Dawei Yin, and Chao Huang.
\newblock Graphgpt: Graph instruction tuning for large language models.
\newblock In \emph{SIGIR}, pp.\  491--500, 2024{\natexlab{a}}.

\bibitem[Tang et~al.(2024{\natexlab{b}})Tang, Yang, Wei, Shi, Xia, Yin, and Huang]{tang2024higpt}
Jiabin Tang, Yuhao Yang, Wei Wei, Lei Shi, Long Xia, Dawei Yin, and Chao Huang.
\newblock Higpt: Heterogeneous graph language model.
\newblock In \emph{KDD}, 2024{\natexlab{b}}.

\bibitem[Veli{\v{c}}kovi{\'c} et~al.(2018)Veli{\v{c}}kovi{\'c}, Cucurull, Casanova, Romero, Lio, and Bengio]{velivckovic2017graph}
Petar Veli{\v{c}}kovi{\'c}, Guillem Cucurull, Arantxa Casanova, Adriana Romero, Pietro Lio, and Yoshua Bengio.
\newblock Graph attention networks.
\newblock In \emph{ICLR}, 2018.

\bibitem[Velickovic et~al.(2018)Velickovic, Cucurull, Casanova, Romero, et~al.]{gat}
Petar Velickovic, Guillem Cucurull, Arantxa Casanova, Adriana Romero, et~al.
\newblock Graph attention networks.
\newblock In \emph{{ICLR} (Poster)}. OpenReview.net, 2018.

\bibitem[Wang et~al.(2019{\natexlab{a}})Wang, Ji, Shi, Wang, Ye, Cui, and Yu]{wang2019heterogeneous}
Xiao Wang, Houye Ji, Chuan Shi, Bai Wang, Yanfang Ye, Peng Cui, and Philip~S Yu.
\newblock Heterogeneous graph attention network.
\newblock In \emph{WWW}, pp.\  2022--2032, 2019{\natexlab{a}}.

\bibitem[Wang et~al.(2019{\natexlab{b}})Wang, Ji, Shi, Wang, Ye, et~al.]{han}
Xiao Wang, Houye Ji, Chuan Shi, Bai Wang, Yanfang Ye, et~al.
\newblock Heterogeneous graph attention network.
\newblock In \emph{{WWW}}, pp.\  2022--2032. {ACM}, 2019{\natexlab{b}}.

\bibitem[Wang et~al.(2022)Wang, Bo, Shi, Fan, Ye, and Philip]{wang2022survey}
Xiao Wang, Deyu Bo, Chuan Shi, Shaohua Fan, Yanfang Ye, and S~Yu Philip.
\newblock A survey on heterogeneous graph embedding: methods, techniques, applications and sources.
\newblock \emph{TBD}, 9\penalty0 (2):\penalty0 415--436, 2022.

\bibitem[Wei et~al.(2022)Wei, Wang, Schuurmans, Bosma, Xia, Chi, Le, Zhou, et~al.]{wei2022chain}
Jason Wei, Xuezhi Wang, Dale Schuurmans, Maarten Bosma, Fei Xia, Ed~Chi, Quoc~V Le, Denny Zhou, et~al.
\newblock Chain-of-thought prompting elicits reasoning in large language models.
\newblock \emph{NeurIPS}, 35:\penalty0 24824--24837, 2022.

\bibitem[Wu et~al.(2024)Wu, Zhang, and Fan]{wu2024graph}
Zhihao Wu, Zhao Zhang, and Jicong Fan.
\newblock Graph convolutional kernel machine versus graph convolutional networks.
\newblock \emph{NeurIPS}, 36, 2024.

\bibitem[Wu et~al.(2020)Wu, Pan, Chen, Long, Zhang, and Philip]{wu2020comprehensive}
Zonghan Wu, Shirui Pan, Fengwen Chen, Guodong Long, Chengqi Zhang, and S~Yu Philip.
\newblock A comprehensive survey on graph neural networks.
\newblock \emph{TPAMI}, 32\penalty0 (1):\penalty0 4--24, 2020.

\bibitem[Xia \& Huang(2024)Xia and Huang]{xia2024anygraph}
Lianghao Xia and Chao Huang.
\newblock Anygraph: Graph foundation model in the wild.
\newblock \emph{arXiv preprint arXiv:2408.10700}, 2024.

\bibitem[Xie et~al.(2023)Xie, Zhou, Cheng, Shi, Weng, Liu, Hua, Zhao, Liu, Liu, et~al.]{xie2023openagents}
Tianbao Xie, Fan Zhou, Zhoujun Cheng, Peng Shi, Luoxuan Weng, Yitao Liu, Toh~Jing Hua, Junning Zhao, Qian Liu, Che Liu, et~al.
\newblock Openagents: An open platform for language agents in the wild.
\newblock In \emph{COLM}, 2023.

\bibitem[Xu et~al.(2020)Xu, Zhang, Mao, Wang, Xie, and Zhang]{xu2020curriculum}
Benfeng Xu, Licheng Zhang, Zhendong Mao, Quan Wang, Hongtao Xie, and Yongdong Zhang.
\newblock Curriculum learning for natural language understanding.
\newblock In \emph{ACL}, pp.\  6095--6104, 2020.

\bibitem[Yang et~al.(2024)Yang, Yang, Hui, Zheng, Yu, Zhou, Li, Li, Liu, Huang, et~al.]{yang2024qwen2}
An~Yang, Baosong Yang, Binyuan Hui, Bo~Zheng, Bowen Yu, Chang Zhou, Chengpeng Li, Chengyuan Li, Dayiheng Liu, Fei Huang, et~al.
\newblock Qwen2 technical report.
\newblock \emph{arXiv preprint arXiv:2407.10671}, 2024.

\bibitem[Yang et~al.(2020)Yang, Ding, Zhou, Yang, Zhou, and Tang]{yang2020understanding}
Zhen Yang, Ming Ding, Chang Zhou, Hongxia Yang, Jingren Zhou, and Jie Tang.
\newblock Understanding negative sampling in graph representation learning.
\newblock In \emph{KDD}, pp.\  1666--1676, 2020.

\bibitem[Yao et~al.(2023)Yao, Zhao, Yu, Du, Shafran, Narasimhan, and Cao]{yaoreact}
Shunyu Yao, Jeffrey Zhao, Dian Yu, Nan Du, Izhak Shafran, Karthik~R Narasimhan, and Yuan Cao.
\newblock React: Synergizing reasoning and acting in language models.
\newblock In \emph{ICLR}, 2023.

\bibitem[Zhang et~al.(2019)Zhang, Song, Huang, Swami, and Chawla]{hetgnn}
Chuxu Zhang, Dongjin Song, Chao Huang, Ananthram Swami, and Nitesh~V. Chawla.
\newblock Heterogeneous graph neural network.
\newblock In \emph{{KDD}}, pp.\  793--803. {ACM}, 2019.

\bibitem[Zhang et~al.(2022)Zhang, Yin, Sheng, Li, Ouyang, Li, Tao, Yang, and Cui]{zhang2022graph}
Wentao Zhang, Ziqi Yin, Zeang Sheng, Yang Li, Wen Ouyang, Xiaosen Li, Yangyu Tao, Zhi Yang, and Bin Cui.
\newblock Graph attention multi-layer perceptron.
\newblock In \emph{KDD}, pp.\  4560--4570, 2022.

\bibitem[Zheng et~al.(2024)Zheng, Chiang, Sheng, Zhuang, Wu, Zhuang, Lin, Li, Li, Xing, et~al.]{zheng2024judging}
Lianmin Zheng, Wei-Lin Chiang, Ying Sheng, Siyuan Zhuang, Zhanghao Wu, Yonghao Zhuang, Zi~Lin, Zhuohan Li, Dacheng Li, Eric Xing, et~al.
\newblock Judging llm-as-a-judge with mt-bench and chatbot arena.
\newblock \emph{NeurIPS}, 36, 2024.

\bibitem[Zhong \& Mottin(2023)Zhong and Mottin]{zhong2023knowledge}
Zhiqiang Zhong and Davide Mottin.
\newblock Knowledge-augmented graph machine learning for drug discovery: From precision to interpretability.
\newblock In \emph{KDD}, pp.\  5841--5842, 2023.

\end{thebibliography}
